\documentclass[11pt]{article}

\usepackage[preprint]{acl}
\usepackage{newfloat}
\usepackage{listings}
\usepackage{float}
\DeclareCaptionStyle{ruled}{labelfont=normalfont,labelsep=colon,strut=off} 
\floatstyle{ruled}
\newfloat{listing}{tb}{lst}{}
\floatname{listing}{Listing}
\usepackage{amsfonts}
\usepackage{amsthm}
\newtheorem{proposition}{Proposition}
\usepackage{bm}
\usepackage{enumitem}
\usepackage{booktabs}
\usepackage[most]{tcolorbox}
\usepackage{multicol}
\lstdefinestyle{promptjson}{
  basicstyle=\ttfamily\footnotesize,
  breaklines=true,
  breakatwhitespace=true,
  columns=fullflexible,
  frame=none
}

\newtcolorbox{PromptBox}[1][]{
  enhanced,
  breakable,
  colback=gray!4,
  colframe=gray!60,
  boxrule=0.6pt,
  arc=2pt,
  left=6pt,right=6pt,top=6pt,bottom=6pt,
  fonttitle=\bfseries,
  title={Prompt Example},
  #1
}
\usepackage{times}
\usepackage{latexsym}
\usepackage{amsmath}
\usepackage{booktabs}
\usepackage[T1]{fontenc}

\usepackage[utf8]{inputenc}

\usepackage{microtype}

\usepackage{inconsolata}

\usepackage{graphicx}

\title{HERO: Hierarchical rationalE Reasoning Optimization for Radiology Report Generation via Reason-then-Summarize}

\author{ 
	Kun Zhao\textsuperscript{1},
	Guodong Liu\textsuperscript{2},
	Hui Ji\textsuperscript{1},
	Siyuan Dai\textsuperscript{1},
	Pan Wang\textsuperscript{1},
	Jifeng Song\textsuperscript{1},\\ 
	\textbf{Heng Huang}\textsuperscript{3},
	\textbf{Chenghua Lin}\textsuperscript{5},
    \textbf{Liang Zhan}\textsuperscript{1},
    \textbf{Haoteng Tang}\textsuperscript{4}\space\\ 
	\textsuperscript{1}University of Pittsburgh\vspace{-0.5mm} \space
	\textsuperscript{2}Eli and Lilly Company\vspace{-0.5mm} \space
	\textsuperscript{3}University of Maryland College Park\vspace{1mm} \\
	\textsuperscript{4}University of Texas Rio Grande Valley\vspace{-0.5mm} \space
    \textsuperscript{5}The University of Manchester\vspace{-0.5mm} \\
	\texttt{
		\{kun.zhao, liang.zhan\}@pitt.edu, 
	}\vspace{-0.5mm} 
	\texttt{
		haoteng.tang@utrgv.edu 
	}\vspace{-0.5mm} \\
	\vspace{-0.5mm}
	\texttt{
		chenghua.lin@manchester.ac.uk
	}\vspace{-0.5mm}
}

\begin{document}
\maketitle
\begin{abstract}

Multimodal Large Language Models (MLLMs) have substantially advanced Radiology Report Generation (RRG), yet aligning them through reinforcement learning (RL) remains challenging due to heterogeneous medical supervision. Vanilla Group Relative Policy Optimization (GRPO) assigns uniform credit across the entire generation, leading to segment interference, token dilution, and rationale--diagnosis decoupling, which exacerbates clinical hallucinations. We propose \textbf{HERO} (\emph{Hierarchical rationalE Reasoning Optimization}), a factorized policy optimization framework that aligns heterogeneous supervision with three optimization granularities. HERO separately optimizes reasoning, diagnosis, and rationale faithfulness through complementary segment-, token-, and completion-level optimization with a heterogeneous reward formulation covering diagnostic accuracy, reasoning quality, and think--answer consistency. Experiments on MIMIC-CXR and IU-Xray show that HERO outperforms strong supervised and reinforcement learning baselines, achieving state-of-the-art clinical efficacy while producing more rationale-supported and think--answer-consistent reports, thereby substantially mitigating clinical hallucinations.

\end{abstract}
\section{Introduction}
Multimodal Large Language Models (MLLMs) have achieved remarkable progress in vision--language learning \citep{bai2025qwen2, li2023llava, wang2025models}, demonstrating strong performance in image captioning \citep{rotstein2024fusecap, chen2024chexagent}, visual question answering \citep{xu2024mlevlm, kuang2025natural}, and medical image understanding \citep{liu2025application, wang2024large, zhao2024x, zhao2025r, gu2025adapting}.
Radiology Report Generation (RRG), however, remains particularly challenging because it requires accurate visual grounding and consistency between radiological findings and diagnostic conclusions \citep{Yi2025ASO}.
RRG aims to generate free-text radiology reports from chest X-rays, reducing radiologists' workload while maintaining diagnostic quality.

Despite recent advances \citep{chen2024chexagent, hein2025chexalign, zambrano2025clinically}, the reliability of RRG remains a major barrier to real-world deployment. Standard supervised fine-tuning (SFT) often relies on linguistic priors rather than image-grounded findings, producing clinically plausible but hallucinated reports
\citep{liu2025rrg, chen2024detecting, wang-etal-2025-semantic,
heiman2025factchexcker}. Reinforcement learning (RL) instead directly optimizes clinically meaningful objectives beyond surface-level text similarity, which correlates only weakly with diagnostic correctness
\citep{yu2023evaluating, ostmeier2024green}. Among existing RL methods, PPO requires a critic network
\citep{zhou2024large, yang2025aligning}, whereas DPO enables efficient offline optimization but lacks online exploration and iterative policy improvement
\citep{liu2025rrg, hein2025chexalign}. GRPO
\citep{shao2024deepseekmath} bridges this gap through critic-free online optimization, making it a compelling foundation for RRG.

Casting RRG into a \emph{think--answer} format makes GRPO applicable, yet vanilla GRPO remains fundamentally mismatched because medical supervision is inherently \emph{heterogeneous}.
A single radiology rollout contains three distinct learning signals---reasoning quality, diagnostic correctness, and consistency between them---each requiring different optimization behaviors. Vanilla GRPO instead assigns a single scalar advantage to the entire completion and broadcasts it uniformly across all tokens, resulting in coarse credit assignment
\citep{tan2025gtpo, guo2026segment}. This mismatch leads to three optimization failures.
\textbf{(1) Segment interference.}
The think and answer segments optimize different objectives---open-ended reasoning versus calibrated diagnosis---yet receive identical gradients, causing supervision to interfere across segments.
\textbf{(2) Token dilution.}
Diagnostic tokens (e.g., `cardiomegaly'') receive the same weight as template tokens (e.g., `the''), wasting gradient budget on uninformative positions.
\textbf{(3) Rationale--diagnosis decoupling.}
GRPO does not explicitly enforce that diagnostic conclusions be supported by the generated reasoning. Consequently, a model may produce unsupported diagnoses or fabricate plausible rationales, consistent with the broader observation that chain-of-thought does not necessarily determine the answer
\citep{turpin2023language, chen2025reasoning}. Such decoupling is particularly problematic in clinical settings because it disconnects the report's supporting rationale from its diagnostic conclusion
\citep{lai2025med, pan2025medvlm}.

These failures share a common cause and call for a unified solution: heterogeneous supervision should be routed to the optimization granularity it governs, rather than collapsed into a uniformly broadcast scalar reward.
Existing methods address different manifestations of this problem, each focusing on a single optimization granularity.
Token-level methods such as DiTPO/DEER \citep{lu2026rethinking} and GTPO \citep{tan2025gtpo} improve token credit assignment, whereas segment- and process-level methods optimize structured reasoning \citep{guo2026segment, mei2026fir, khalifa2025process}.
Rationale- and consistency-aware methods, including ESC-RL \citep{zhou2026enhancing} and preference-based RRG \citep{liu2025rrg, hein2025chexalign}, encourage consistency under external clinical supervision but do not explicitly optimize the model's internal think--answer consistency.
More fundamentally, this supervision is still compressed into a single scalar advantage during policy optimization.
In RRG, reasoning, diagnostic, and consistency supervision naturally align with the segment, token, and completion levels, motivating our factorized optimization framework.

In this work, we propose \textbf{HERO} (\emph{Hierarchical Rationale Reasoning Optimization}), a factorized policy optimization framework that routes heterogeneous supervision to the optimization granularity where it is most informative.
HERO factorizes GRPO into three complementary optimization factors.
A \textbf{segment} factor separately optimizes reasoning and diagnosis.
A \textbf{token} factor emphasizes diagnostically informative tokens via salience-aware reweighting.
A \textbf{rationale-supported} completion factor encourages diagnostic conclusions to be supported by the generated reasoning, discouraging hallucinated rationales.
A heterogeneous reward further decomposes report quality into diagnostic accuracy, reasoning quality, and think--answer consistency, providing targeted supervision for each optimization factor.
Empirically, HERO achieves state-of-the-art clinical performance on MIMIC-CXR (e.g., RadGraph $0.354$, Micro-F1(14) $0.574$), outperforming MPO, ESC-RL, and other RL baselines while generalizing to IU-Xray. Moreover, HERO is the only system that is simultaneously think--answer consistent and rationale-faithful (rationale faithfulness score $0.773$ versus $\approx0.65$ for strong VLM baselines), directly validating the proposed rationale-supported optimization.

Our contributions are summarized as follows:
\begin{itemize}
\item We formulate heterogeneous supervision in radiology report generation as a factorized policy optimization problem and propose HERO, which routes supervision to its corresponding optimization granularity.
\item HERO jointly optimizes reasoning quality, diagnostic accuracy, and think--answer consistency through a factorized GRPO objective with complementary segment-, token-, and completion-level optimization factors and a heterogeneous reward formulation.
\item HERO consistently outperforms state-of-the-art RL baselines on MIMIC-CXR, generalizes to IU-Xray, and substantially improves rationale-supported reasoning, reducing clinical hallucinations while maintaining strong diagnostic performance.
\end{itemize}

\begin{figure*}[h]
  \centering
  \includegraphics[width=0.8\linewidth]{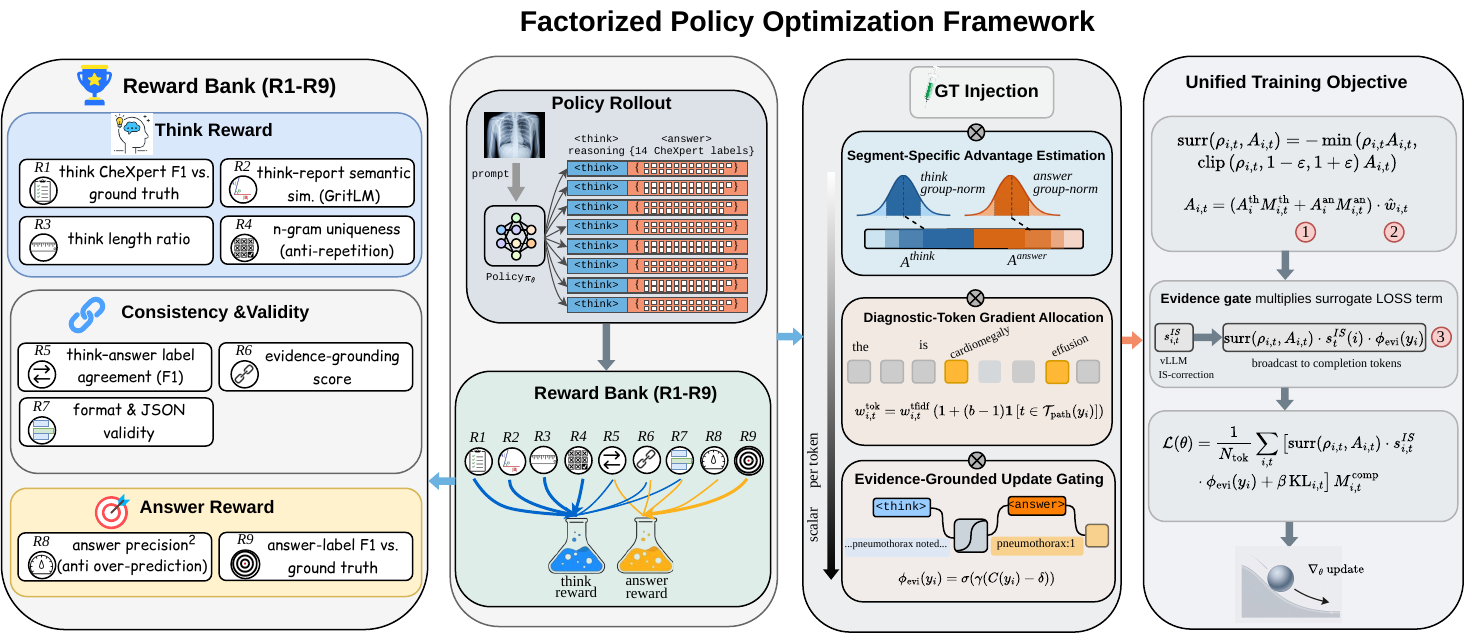}
    \caption{Overview of the HERO framework. During policy rollout, completions are generated as \texttt{<think>} and \texttt{<answer>} segments, with conditional GT injection to break exploration deadlocks. The model is evaluated by a heterogeneous Reward Bank ($R_1$-$R_9$). To optimize the policy, HERO factorizes the GRPO objective across three granularities: (1) Segment-level: routes rewards to estimate separate advantages for reasoning and diagnosis; (2) Token-level: allocates larger gradient weights to diagnostically salient tokens; and (3) Completion-level: applies an rationale-supported gate to scale the surrogate loss based on think-answer consistency. These factors are jointly optimized in a unified training objective.}
   \label{fig:RDDM}
\end{figure*}


\section{Related Work}
\label{sec:related}

\paragraph{Reinforcement learning for report generation.}
RL aligns generation with clinical objectives beyond token-level likelihood. 
PPO with clinical-quality rewards improves diagnostic metrics on MIMIC-CXR
\citep{zhou2024large} but requires a critic; preference-based methods
\citep{liu2025rrg, hein2025chexalign} are efficient but operate offline; and critic-free GRPO \citep{shao2024deepseekmath} enables online exploration. Closest to our setting, ESC-RL \citep{zhou2026enhancing} combines an evidence-aware reward with self-correcting preference learning. 
These methods improve the \emph{reward}, whereas HERO restructures \emph{credit assignment} within GRPO, making the two approaches complementary.

\paragraph{Credit assignment and reasoning faithfulness.}
A parallel line of RL research for LLMs argues that the uniform, sequence-level advantage of GRPO/DAPO is too coarse: segment-level \citep{guo2026segment} and token-level \citep{tan2025gtpo, lu2026rethinking} methods assign credit at finer granularity and improve reasoning accuracy, while process rewards evaluate intermediate reasoning steps \citep{khalifa2025process, mei2026fir}. 
Separately, faithfulness studies show that a model's stated chain-of-thought often does not causally determine its answer \citep{turpin2023language, lanham2023measuring, chen2025reasoning}, a phenomenon that also persists in medical VLMs \citep{lai2025med, pan2025medvlm}. 
HERO unifies these lines of work by factorizing GRPO credit at the \emph{segment}, \emph{token}, and \emph{completion} levels within a single objective and---unlike consistency-only signals---gating updates on whether the diagnosis is grounded in the ground truth, directly addressing the rationale--diagnosis decoupling identified by these faithfulness studies.

\section{Methods}
\label{sec:method}
\subsection{Problem Formulation}
\label{sec:prob}
Let $\mathcal{D}=\{(x^{(i)},y^{(i)}_{\text{text}},y^{(i)}_{\text{label}})\}_{i=1}^N$ be a
dataset of radiology studies, where $x^{(i)}$ is a chest X-ray,
$y^{(i)}_{\text{text}}$ denotes the free-text findings, and
$y^{(i)}_{\text{label}}$ the structured diagnostic label vector
(the fourteen CheXpert pathologies). 
We learn a policy
$\pi_\theta$ parameterized by a multimodal LLM that generates a token sequence $Y$ structured into
two segments: a \texttt{<think>} block of free-text radiological observations and an
\texttt{<answer>} block of structured disease labels in JSON. 
The generation process factorizes as
\begin{equation}
P(Y|x)=P(Y_{\text{think}}|x)\;P(Y_{\text{answer}}|x,Y_{\text{think}}),
\label{eq:factor}
\end{equation}
where the model first generates free-text findings and then summarizes
them into structured diagnostic labels. 
This reason-then-summarize formulation naturally exposes heterogeneous supervision over reasoning quality, diagnostic correctness, and think--answer consistency, motivating a factorized policy optimization framework.

\subsection{Heterogeneous Supervision}
\label{sec:reward}
The think-answer rollout naturally exposes three complementary forms of supervision: \emph{reasoning quality}, \emph{diagnostic correctness}, and \emph{reasoning--diagnosis consistency}. 
Reasoning quality evaluates whether the generated rationale is clinically correct, diagnostic correctness evaluates whether the final diagnosis is correct, and reasoning--diagnosis consistency evaluates whether the diagnosis is faithfully supported by the generated rationale.
We instantiate these supervision signals using nine reward components (Reward Bank in Figure~\ref{fig:RDDM}), each normalized to $[0,1]$ and assigned a weight $w_k$.
The rewards are organized into three groups corresponding to the supervision they quantify: reasoning quality ($R_1$--$R_4$), reasoning--diagnosis consistency and validity ($R_5$--$R_7$), and diagnostic correctness ($R_8$--$R_9$). 
The heterogeneous nature of these supervision signals motivates a factorized policy optimization framework that aligns different forms of supervision with appropriate optimization granularity.

\subsection{Factorized Policy Optimization Framework}
\label{sec:fpo}
HERO is a \emph{factorized policy optimization framework} that aligns heterogeneous supervision with the optimization granularity at which each supervision signal is most informative (Figure \ref{fig:RDDM}). 
Rather than optimizing all supervision signals through a single completion-level scalar advantage, HERO decomposes the policy optimization objective into three complementary optimization factors: a \textbf{segment-specific advantage} that allocates credit between reasoning and diagnosis, a \textbf{diagnostic-token gradient allocation} mechanism that emphasizes clinically informative tokens, and a \textbf{rationale-supported update gate} that regulates completion-level policy updates. 

The following three sections derive our segment-, token-, and completion-level optimization factors from this objective and finally combine them into a unified training objective.

\subsubsection{Segment-Specific Advantage Estimation}
\label{sec:seg}
For each image $x$, we first sample a group of $G$ completions $\{y_i\}_{i=1}^{G}$ from the policy model.
To assign credit separately to reasoning and diagnosis, we route the
reward components into two segment-specific composites:
\begin{equation}
    r_i^{\text{th}} =\sum_{k=1}^{4} w_k R_k \;+\; \lambda\!\sum_{k=5}^{7} w_k R_k,\quad r_i^{\text{an}} =\sum_{k=5}^{9} w_k R_k.
\label{eq:segrew}
\end{equation}

Here, the reasoning-oriented components $R_1$--$R_4$ contribute only to the
\texttt{think} composite, while the answer-specific diagnostic components $R_8$ and $R_9$ contribute only to the \texttt{answer} composite. 
The shared components $R_5$--$R_7$, which measure think--answer agreement, rationale faithfulness, and output validity, contribute fully to the answer composite and to the think composite with a reduced coefficient $\lambda$. 
This design provides additional within-group variation for estimating the reasoning-side advantage, since reasoning-oriented rewards alone often exhibit limited variation across completions sampled for the same image, resulting in weak or unstable standardized advantages. 
To penalize severe formatting errors, any completion that cannot be successfully parsed into valid reasoning and diagnosis segments receives zero reward across both composites, i.e., $r_{i}^{th}=r_{i}^{an}=0$.

These two rewards (i.e., $r_{i}^{th}$ and $r_{i}^{an}$) are independently standardized within
each sampled group:
\begin{equation}
A_i^{s}
=
\frac{r_i^{s}-\bar r^{s}}
{\max(\sigma^{s},\epsilon_{\text{std}})},
\qquad
s\in\{\mathrm{th},\mathrm{an}\},
\label{eq:segnorm}
\end{equation}
where $\bar r^{s}$ and $\sigma^{s}$ denote the group mean and standard
deviation of the corresponding segment reward.

We then define binary masks $M_{i,t}^{\mathrm{th}},M_{i,t}^{\mathrm{an}}\in\{0,1\}$ to identify $t$-th tokens in $i$-th completion (i.e., $y_{i,t}$), belonging to the \texttt{think} and
\texttt{answer} segments, respectively.

We further define
$M_{i,t}^{\mathrm{comp}}\in\{0,1\}$
as the mask over all generated (non-prompt and non-padding) tokens, such that
$M_{i,t}^{\mathrm{th}}+M_{i,t}^{\mathrm{an}}
=
M_{i,t}^{\mathrm{comp}}$.

The segment-specific token advantage can be formulated as:
\begin{equation}
A_{i,t}^{\mathrm{seg}}
=
A_i^{\mathrm{th}}M_{i,t}^{\mathrm{th}}
+
A_i^{\mathrm{an}}M_{i,t}^{\mathrm{an}} .
\label{eq:segadv}
\end{equation}

\subsubsection{Diagnostic-Token Gradient Allocation}
\label{sec:token}
Within each completion, only a subset of tokens carry diagnostically meaningful information, while many others are generic function or structural tokens.
We therefore assign each generated token a diagnostic salience weight
based on TF--IDF statistics computed from the training corpus:
\begin{equation}
w_{i,t}^{\mathrm{tfidf}}
=
(1-\alpha)
+
\alpha
\frac{\mathrm{tfidf}(v_{i,t})}
{\max_{v}\mathrm{tfidf}(v)}
\in [1-\alpha,1],
\label{eq:tfidf}
\end{equation}
where $v_{i,t}$ denotes the word corresponding to token
$y_{i,t}$, $\mathrm{tfidf}(v)$ is its corpus-level TF--IDF score
computed from the training corpus, and $\alpha\in[0,1]$ controls the
strength of the salience adjustment. 
Consequently, rare diagnostic terms
receive weights closer to $1$, whereas common boilerplate terms receive
weights closer to $1-\alpha$.

We further apply a multiplicative boost $b$ to tokens that begin a
predefined pathology keyword:
\begin{equation}
w_{i,t}^{\mathrm{tok}}
=
w_{i,t}^{\mathrm{tfidf}}
\left(
1+(b-1)\mathbf{1}
[t\in\mathcal{T}_{\mathrm{path}}(y_{i})]
\right),
\label{eq:pathboost}
\end{equation}
where $\mathbf{1}[\cdot]$ is the indicator function and
$\mathcal{T}_{\mathrm{path}}(y_i)$ denotes the token positions in
completion $y_i$ that begin a pathology keyword.

Directly applying these weights would alter the total gradient mass of
each completion, introducing an implicit sample-dependent learning-rate
effect. To separate token emphasis from the overall update scale, we
normalize the weights to have mean one over the generated tokens:
\begin{equation}
\hat w_{i,t}
=
w_{i,t}^{\mathrm{tok}}
\cdot
\frac{\sum_{t'}M_{i,t'}^{\mathrm{comp}}}
{\sum_{t'}w_{i,t'}^{\mathrm{tok}}
M_{i,t'}^{\mathrm{comp}}}.
\label{eq:gsn}
\end{equation}
This normalization preserves the total masked weight mass:
\begin{equation}
\sum_t
\hat w_{i,t}M_{i,t}^{\mathrm{comp}}
=
\sum_t M_{i,t}^{\mathrm{comp}}.
\end{equation}
The token factor therefore redistributes gradient emphasis toward
diagnostically informative positions without changing the overall
per-completion scale.

Combining the token weights with the segment-specific advantage yields
the final per-token advantage:
\begin{equation}
A_{i,t}
=
A_{i,t}^{\mathrm{seg}}\hat w_{i,t}.
\label{eq:l1l2}
\end{equation}

\begin{table*}[t]
\small
\centering
\resizebox{1.0\textwidth}{!}{%
\begin{tabular}{l|cccclcccclllccl}
\toprule[1.5pt]
                    & \multicolumn{15}{c}{MIMIC-CXR} \\ \midrule
                    & \multicolumn{5}{c|}{NLG Metrics} & \multicolumn{10}{c}{Clinical Efficacy} \\ \midrule
                    & \multicolumn{1}{c|}{B-4} & \multicolumn{1}{c|}{MTR} & \multicolumn{1}{c|}{RG-L} & \multicolumn{1}{c|}{BERTScore} & \multicolumn{1}{c|}{SRR-BERT} & \multicolumn{1}{c|}{RadGraph} & \multicolumn{1}{c|}{RadCliQ $\downarrow$} & \multicolumn{1}{c|}{RateScore} & \multicolumn{1}{c|}{GREEN} & \multicolumn{1}{c|}{Micro-F1(14)} & \multicolumn{1}{c|}{avg-Macro-F1(14)} & \multicolumn{1}{c|}{w-Macro-F1(14)} & \multicolumn{1}{c|}{Micro-F1(5)} & \multicolumn{1}{c|}{avg-Macro-F1(5)} & \multicolumn{1}{c}{w-Macro-F1(5)} \\ \midrule
Llava-1.5           & \multicolumn{1}{c|}{0.102} & \multicolumn{1}{c|}{0.323} & \multicolumn{1}{c|}{0.353} & \multicolumn{1}{c|}{0.611} & \multicolumn{1}{c|}{\underline{0.501}} & \multicolumn{1}{c|}{\underline{0.320}} & \multicolumn{1}{c|}{1.358} & \multicolumn{1}{c|}{0.637} & \multicolumn{1}{c|}{0.498} & \multicolumn{1}{c|}{0.550} & \multicolumn{1}{c|}{0.408} & \multicolumn{1}{c|}{\textbf{0.533}} & \multicolumn{1}{c|}{\underline{0.593}} & \multicolumn{1}{c|}{0.516} & \multicolumn{1}{c}{\underline{0.583}} \\
Llava-Next(vicuna)     & \multicolumn{1}{c|}{0.054} & \multicolumn{1}{c|}{0.165} & \multicolumn{1}{c|}{0.263} & \multicolumn{1}{c|}{0.438} & \multicolumn{1}{c|}{0.463} & \multicolumn{1}{c|}{0.265} & \multicolumn{1}{c|}{1.245} & \multicolumn{1}{c|}{0.617} & \multicolumn{1}{c|}{0.503} & \multicolumn{1}{c|}{0.408} & \multicolumn{1}{c|}{0.330} & \multicolumn{1}{c|}{0.413} & \multicolumn{1}{c|}{0.419} & \multicolumn{1}{c|}{0.392} & \multicolumn{1}{c}{0.445} \\
Llava-Next(mistral) & \multicolumn{1}{c|}{0.102} & \multicolumn{1}{c|}{\underline{0.325}} & \multicolumn{1}{c|}{0.350} & \multicolumn{1}{c|}{0.609} & \multicolumn{1}{c|}{\textbf{0.505}} & \multicolumn{1}{c|}{0.314} & \multicolumn{1}{c|}{1.326} & \multicolumn{1}{c|}{0.636} & \multicolumn{1}{c|}{0.493} & \multicolumn{1}{c|}{0.544} & \multicolumn{1}{c|}{0.412} & \multicolumn{1}{c|}{0.525} & \multicolumn{1}{c|}{0.574} & \multicolumn{1}{c|}{0.512} & \multicolumn{1}{c}{0.567} \\
Llava-Rad           & \multicolumn{1}{c|}{\underline{0.104}} & \multicolumn{1}{c|}{0.322} & \multicolumn{1}{c|}{0.357} & \multicolumn{1}{c|}{\underline{0.612}} & \multicolumn{1}{c|}{0.494} & \multicolumn{1}{c|}{\underline{0.320}} & \multicolumn{1}{c|}{1.369} & \multicolumn{1}{c|}{\underline{0.640}} & \multicolumn{1}{c|}{0.495} & \multicolumn{1}{c|}{\underline{0.553}} & \multicolumn{1}{c|}{0.411} & \multicolumn{1}{c|}{0.527} & \multicolumn{1}{c|}{0.588} & \multicolumn{1}{c|}{0.512} & \multicolumn{1}{c}{0.576} \\
Qwen-2.5-VL-7B        & \multicolumn{1}{c|}{\textbf{0.110}} & \multicolumn{1}{c|}{0.317} & \multicolumn{1}{c|}{0.348} & \multicolumn{1}{c|}{0.604} & \multicolumn{1}{c|}{0.473} & \multicolumn{1}{c|}{0.308} & \multicolumn{1}{c|}{1.274} & \multicolumn{1}{c|}{0.627} & \multicolumn{1}{c|}{0.482} & \multicolumn{1}{c|}{0.519} & \multicolumn{1}{c|}{0.377} & \multicolumn{1}{c|}{0.495} & \multicolumn{1}{c|}{0.554} & \multicolumn{1}{c|}{0.481} & \multicolumn{1}{c}{0.544} \\ 
\midrule
R2Gen               & \multicolumn{1}{c|}{0.078} & \multicolumn{1}{c|}{0.309} & \multicolumn{1}{c|}{\textbf{0.366}} & \multicolumn{1}{c|}{0.604} & \multicolumn{1}{c|}{0.435} & \multicolumn{1}{c|}{0.318} & \multicolumn{1}{c|}{1.203} & \multicolumn{1}{c|}{0.608} & \multicolumn{1}{c|}{0.474} & \multicolumn{1}{c|}{0.476} & \multicolumn{1}{c|}{0.341} & \multicolumn{1}{c|}{0.437} & \multicolumn{1}{c|}{0.516} & \multicolumn{1}{c|}{0.443} & \multicolumn{1}{c}{0.500} \\
R2GenCMN            & \multicolumn{1}{c|}{0.079} & \multicolumn{1}{c|}{0.309} & \multicolumn{1}{c|}{\underline{0.364}} & \multicolumn{1}{c|}{0.603} & \multicolumn{1}{c|}{0.434} & \multicolumn{1}{c|}{0.317} & \multicolumn{1}{c|}{1.161} & \multicolumn{1}{c|}{0.609} & \multicolumn{1}{c|}{0.469} & \multicolumn{1}{c|}{0.465} & \multicolumn{1}{c|}{0.338} & \multicolumn{1}{c|}{0.438} & \multicolumn{1}{c|}{0.480} & \multicolumn{1}{c|}{0.418} & \multicolumn{1}{c}{0.467} \\
MPO                 & \multicolumn{1}{c|}{0.092} & \multicolumn{1}{c|}{0.308} & \multicolumn{1}{c|}{0.340} & \multicolumn{1}{c|}{0.602} & \multicolumn{1}{c|}{0.471} & \multicolumn{1}{c|}{0.311} & \multicolumn{1}{c|}{1.268} & \multicolumn{1}{c|}{0.624} & \multicolumn{1}{c|}{0.481} & \multicolumn{1}{c|}{0.516} & \multicolumn{1}{c|}{0.370} & \multicolumn{1}{c|}{0.486} & \multicolumn{1}{c|}{0.559} & \multicolumn{1}{c|}{0.491} & \multicolumn{1}{c}{0.550} \\
ESC RL              & \multicolumn{1}{c|}{0.100} & \multicolumn{1}{c|}{0.317} & \multicolumn{1}{c|}{0.350} & \multicolumn{1}{c|}{0.607} & \multicolumn{1}{c|}{0.480} & \multicolumn{1}{c|}{0.309} & \multicolumn{1}{c|}{1.277} & \multicolumn{1}{c|}{0.627} & \multicolumn{1}{c|}{0.484} & \multicolumn{1}{c|}{0.520} & \multicolumn{1}{c|}{0.386} & \multicolumn{1}{c|}{0.497} & \multicolumn{1}{c|}{0.558} & \multicolumn{1}{c|}{0.478} & \multicolumn{1}{c}{0.545} \\
EditGRPO            & \multicolumn{1}{c|}{0.093} & \multicolumn{1}{c|}{0.314} & \multicolumn{1}{c|}{0.339} & \multicolumn{1}{c|}{0.604} & \multicolumn{1}{c|}{0.472} & \multicolumn{1}{c|}{0.311} & \multicolumn{1}{c|}{1.260} & \multicolumn{1}{c|}{0.622} & \multicolumn{1}{c|}{0.484} & \multicolumn{1}{c|}{0.517} & \multicolumn{1}{c|}{0.383} & \multicolumn{1}{c|}{0.492} & \multicolumn{1}{c|}{0.550} & \multicolumn{1}{c|}{0.476} & \multicolumn{1}{c}{0.539} \\
DPO (GREEN)         & \multicolumn{1}{c|}{0.061} & \multicolumn{1}{c|}{0.243} & \multicolumn{1}{c|}{0.231} & \multicolumn{1}{c|}{0.502} & \multicolumn{1}{c|}{0.481} & \multicolumn{1}{c|}{0.284} & \multicolumn{1}{c|}{\textbf{0.927}} & \multicolumn{1}{c|}{0.609} & \multicolumn{1}{c|}{\textbf{0.568}} & \multicolumn{1}{c|}{0.473} & \multicolumn{1}{c|}{0.338} & \multicolumn{1}{c|}{0.440} & \multicolumn{1}{c|}{0.555} & \multicolumn{1}{c|}{0.486} & \multicolumn{1}{c}{0.537} \\ \midrule
Qwen3.5-122B        & \multicolumn{1}{c|}{0.036} & \multicolumn{1}{c|}{0.267} & \multicolumn{1}{c|}{0.234} & \multicolumn{1}{c|}{0.539} & \multicolumn{1}{c|}{0.437} & \multicolumn{1}{c|}{0.227} & \multicolumn{1}{c|}{1.857} & \multicolumn{1}{c|}{0.575} & \multicolumn{1}{c|}{\underline{0.534}} & \multicolumn{1}{c|}{0.445} & \multicolumn{1}{c|}{0.332} & \multicolumn{1}{c|}{0.429} & \multicolumn{1}{c|}{0.461} & \multicolumn{1}{c|}{0.437} & \multicolumn{1}{c}{0.465} \\
InternVL-241B       & \multicolumn{1}{c|}{0.039} & \multicolumn{1}{c|}{0.253} & \multicolumn{1}{c|}{0.233} & \multicolumn{1}{c|}{0.549} & \multicolumn{1}{c|}{0.435} & \multicolumn{1}{c|}{0.213} & \multicolumn{1}{c|}{1.868} & \multicolumn{1}{c|}{0.593} & \multicolumn{1}{c|}{0.528} & \multicolumn{1}{c|}{0.404} & \multicolumn{1}{c|}{0.322} & \multicolumn{1}{c|}{0.429} & \multicolumn{1}{c|}{0.454} & \multicolumn{1}{c|}{0.453} & \multicolumn{1}{c}{0.486} \\ \midrule
DPO (Our rewards)   & \multicolumn{1}{c|}{0.073} & \multicolumn{1}{c|}{0.273} & \multicolumn{1}{c|}{0.251} & \multicolumn{1}{c|}{0.523} & \multicolumn{1}{c|}{0.478} & \multicolumn{1}{c|}{0.305} & \multicolumn{1}{c|}{\underline{0.996}} & \multicolumn{1}{c|}{0.615} & \multicolumn{1}{c|}{0.451} & \multicolumn{1}{c|}{0.532} & \multicolumn{1}{c|}{\textbf{0.417}} & \multicolumn{1}{c|}{0.529} & \multicolumn{1}{c|}{0.574} & \multicolumn{1}{c|}{\underline{0.518}} & \multicolumn{1}{c}{0.572} \\
Ours               & \multicolumn{1}{c|}{0.102}          & \multicolumn{1}{c|}{\textbf{0.334}}          & \multicolumn{1}{c|}{0.340}          & \multicolumn{1}{c|}{\textbf{0.617}}          & \multicolumn{1}{c|}{0.487}          & \multicolumn{1}{c|}{\textbf{0.354}}          & \multicolumn{1}{c|}{1.178}          & \multicolumn{1}{c|}{\textbf{0.649}}          & \multicolumn{1}{c|}{0.498}          & \multicolumn{1}{c|}{\textbf{0.574}}          & \multicolumn{1}{c|}{\underline{0.413}}          & \multicolumn{1}{c|}{\underline{0.530}}          & \multicolumn{1}{c|}{\textbf{0.618}}          & \multicolumn{1}{c|}{\textbf{0.525}}          & \multicolumn{1}{c}{\textbf{0.589}} \\
\bottomrule[1.5pt]
\end{tabular}%
}
\caption{Report-level MIMIC-CXR results, all methods re-scored under one unified protocol. NLG (left) and clinical-efficacy (right) metrics; RadCliQ ($\downarrow$) is lower-is-better, all others higher-is-better. \textbf{Bold} best, \underline{underline} runner-up per column. \emph{DPO (Our rewards)} is an optimizer-swap ablation under our reward.}
\label{tab:mimic}
\end{table*}
\subsubsection{Rationale-Supported Update Gating}
\label{sec:gate}

The completion-level factor evaluates whether the predicted diagnosis is
supported both by the generated reasoning and by the ground-truth
labels. For each pathology $k$, we extract an affirmed rationale indicator
$\psi_{i,k}\in \{0,1\}$ from the \texttt{think} segment of completion
$y_i$ using a pathology term lexicon and a clause-level negation and
uncertainty detector. A matched pathology mention is assigned
$\psi_{i,k}=0$ when a negation or uncertainty cue, such as ``no,''
``without,'' or ``cannot exclude,'' occurs within the same clause.
Thus, a negated statement such as ``no pneumothorax'' is not treated as
affirmed rationale for pneumothorax.

We combine the rationale indicator $\psi_{i,k}$, the predicted answer
label $a_{i,k}$, and the ground-truth label $g_{i,k}$ through a
predefined scoring matrix that produces a per-pathology faithfulness score
$s_{i,k}$. The scoring matrix rewards correctly evidenced true
positives,
$a_{i,k}=g_{i,k}=\psi_{i,k}=1$, and correctly handled true negatives,
including both absent mentions and explicitly negated findings. It
penalizes fabricated positives,
$a_{i,k}=1$ and $g_{i,k}=0$, as well as missed positives,
$g_{i,k}=1$ and $a_{i,k}=0$. The complete scoring matrix is provided in the Supplementary Material.

Let $\mathcal{P}'_i=\{k:g_{i,k}\notin\{\mathrm{null},-1\}\}$ denote the set of pathologies with evaluable ground-truth labels for completion $y_i$.
The rationale faithfulness score is
\begin{equation}
C(y_i)
=
\frac{1}{|\mathcal{P}'_i|}
\sum_{k\in\mathcal{P}'_i}s_{i,k}
\in[-1,1].
\label{eq:cscore}
\end{equation}

The faithfulness score serves two complementary roles. First, it is linearly
mapped to $[0,1]$ as the reward component $R_6$, contributing to the
group-relative advantages and thus influencing the relative ranking of
sampled completions. Second, it defines a completion-level update gate:
\begin{equation}
\phi_{\mathrm{rat}}(y_i)
=
\sigma\!\left(
\gamma\left(C(y_i)-\delta\right)
\right)
\in(0,1),
\label{eq:gate}
\end{equation}
where $\sigma(\cdot)$ is the logistic sigmoid, $\gamma$ controls the
slope, and $\delta$ specifies the faithfulness threshold. The gate scales
the policy update of the entire completion, so that weakly grounded
completions receive smaller updates.
Accordingly, $R_6$ affects the relative advantage assigned to each
completion, whereas $\phi_{\mathrm{rat}}(y_i)$ modulates the magnitude of
its policy update.




\subsubsection{Unified Training Objective}
\label{sec:obj}

The three optimization factors are combined into a unified
DAPO-style token-level objective. The segment-specific advantage and
diagnostic-token weight determine the per-token advantage, while the
rationale-supported gate controls the update magnitude of the entire
completion:
\begin{equation}
\begin{aligned}
\mathcal{L}(\theta)
=
\frac{1}{N_{\mathrm{tok}}}
\sum_{i,t}
\Big[
&\phi_{\mathrm{rat}}(y_i)\,
s_{i,t}^{\mathrm{IS}}\,
\operatorname{surr}
\big(\rho_{i,t},A_{i,t}\big)
\\
&+\beta\,\mathrm{KL}_{i,t}
\Big]
M_{i,t}^{\mathrm{comp}},
\end{aligned}
\label{eq:compact}
\end{equation}
where $A_{i,t}
=
\underbrace{
\left(
A_i^{\mathrm{th}}M_{i,t}^{\mathrm{th}}
+
A_i^{\mathrm{an}}M_{i,t}^{\mathrm{an}}
\right)
}_{\text{segment factor}}
\underbrace{\hat w_{i,t}}_{\text{token factor}}\nonumber
\label{eq:final_adv}$,
$\phi_{\mathrm{rat}}(y_i)$ is the completion-level factor defined in
Eq.~\eqref{eq:gate}. The normalization term $N_{\mathrm{tok}}$ is the total number of
generated completion tokens. $\mathrm{KL}_{i,t}$ denotes the $k_3$ KL-estimator penalty
between the current policy $\pi_\theta$ and a frozen reference model
$\pi_{\mathrm{ref}}$, with coefficient $\beta$. 

For notational compactness, we omit the conditioning context
$(x,y_{i,<t})$ from the policy probabilities. The per-token policy ratio
is
\begin{equation}
\rho_{i,t}
=
\exp\left(
\log\pi_\theta(y_{i,t})
-
\log\pi_{\mathrm{old}}(y_{i,t})
\right),
\end{equation}
and the clipped surrogate loss is
\begin{equation}
\operatorname{surr}(\rho,A)
=
-\min\left(
\rho A,\,
\operatorname{clip}
(\rho,1-\varepsilon,1+\varepsilon)A
\right),
\end{equation}
where $\varepsilon$ is the clipping radius.

Since completions are generated using the inference-engine policy
$\pi_{\mathrm{gen}}$, we apply a per-token importance-sampling correction
for its mismatch with the behavior policy: $s_{i,t}^{\mathrm{IS}}=\exp\left(\log\pi_{\mathrm{old}}(y_{i,t})-\log\pi_{\mathrm{gen}}(y_{i,t})\right).$

To improve optimization stability, we conditionally inject the ground-truth sequence into a sampled group when its reward statistics indicate degeneracy (e.g., exceptionally small group mean and standard deviation). 
The injected sequence provides a high-quality anchor that restores meaningful reward variation and facilitates subsequent policy updates.

\begin{table*}[t]
\centering
\small
\resizebox{1.0\textwidth}{!}{%
\begin{tabular}{l|cccclcccclllccl}
\toprule[1.5pt]
                    & \multicolumn{15}{c}{IU-Xray} \\ \midrule
                    & \multicolumn{5}{c|}{NLG Metrics} & \multicolumn{10}{c}{Clinical Efficacy} \\ \midrule
                    & \multicolumn{1}{c|}{B-4} & \multicolumn{1}{c|}{MTR} & \multicolumn{1}{c|}{RG-L} & \multicolumn{1}{c|}{BERTScore} & \multicolumn{1}{c|}{SRR-BERT} & \multicolumn{1}{c|}{RadGraph} & \multicolumn{1}{c|}{RadCliQ $\downarrow$} & \multicolumn{1}{c|}{RateScore} & \multicolumn{1}{c|}{GREEN} & \multicolumn{1}{c|}{Micro-F1(14)} & \multicolumn{1}{c|}{avg-Macro-F1(14)} & \multicolumn{1}{c|}{w-Macro-F1(14)} & \multicolumn{1}{c|}{Micro-F1(5)} & \multicolumn{1}{c|}{avg-Macro-F1(5)} & \multicolumn{1}{c}{w-Macro-F1(5)} \\ \midrule
Llava-1.5   & \multicolumn{1}{c|}{0.072} & \multicolumn{1}{c|}{0.293} & \multicolumn{1}{c|}{0.332} & \multicolumn{1}{c|}{0.597} & \multicolumn{1}{c|}{0.483} & \multicolumn{1}{c|}{0.310} & \multicolumn{1}{c|}{1.073} & \multicolumn{1}{c|}{0.613} & \multicolumn{1}{c|}{0.628} & \multicolumn{1}{c|}{0.387} & \multicolumn{1}{c|}{0.076} & \multicolumn{1}{c|}{0.247} & \multicolumn{1}{c|}{0.033} & \multicolumn{1}{c|}{0.019} & \multicolumn{1}{c}{0.030} \\
Llava-Next(vicuna)  & \multicolumn{1}{c|}{0.039} & \multicolumn{1}{c|}{0.151} & \multicolumn{1}{c|}{0.167} & \multicolumn{1}{c|}{0.388} & \multicolumn{1}{c|}{\underline{0.494}} & \multicolumn{1}{c|}{0.315} & \multicolumn{1}{c|}{\textbf{0.647}} & \multicolumn{1}{c|}{0.646} & \multicolumn{1}{c|}{0.622} & \multicolumn{1}{c|}{0.164} & \multicolumn{1}{c|}{0.091} & \multicolumn{1}{c|}{0.217} & \multicolumn{1}{c|}{0.113} & \multicolumn{1}{c|}{0.099} & \multicolumn{1}{c}{0.120} \\
Llava-Next(mistral) & \multicolumn{1}{c|}{0.078} & \multicolumn{1}{c|}{0.294} & \multicolumn{1}{c|}{0.344} & \multicolumn{1}{c|}{0.548} & \multicolumn{1}{c|}{0.463} & \multicolumn{1}{c|}{0.308} & \multicolumn{1}{c|}{1.167} & \multicolumn{1}{c|}{0.577} & \multicolumn{1}{c|}{0.482} & \multicolumn{1}{c|}{0.329} & \multicolumn{1}{c|}{0.047} & \multicolumn{1}{c|}{0.127} & \multicolumn{1}{c|}{0.104} & \multicolumn{1}{c|}{0.012} & \multicolumn{1}{c}{0.078} \\
Llava-Rad           & \multicolumn{1}{c|}{\textbf{0.096}} & \multicolumn{1}{c|}{\underline{0.320}} & \multicolumn{1}{c|}{\textbf{0.365}} & \multicolumn{1}{c|}{\underline{0.612}} & \multicolumn{1}{c|}{\textbf{0.495}} & \multicolumn{1}{c|}{\underline{0.363}} & \multicolumn{1}{c|}{1.462} & \multicolumn{1}{c|}{\underline{0.655}} & \multicolumn{1}{c|}{0.587} & \multicolumn{1}{c|}{\underline{0.388}} & \multicolumn{1}{c|}{0.089} & \multicolumn{1}{c|}{0.260} & \multicolumn{1}{c|}{0.147} & \multicolumn{1}{c|}{0.100} & \multicolumn{1}{c}{0.139} \\
Qwen-2.5-VL-7B        & \multicolumn{1}{c|}{0.086} & \multicolumn{1}{c|}{0.293} & \multicolumn{1}{c|}{0.335} & \multicolumn{1}{c|}{0.596} & \multicolumn{1}{c|}{0.486} & \multicolumn{1}{c|}{0.313} & \multicolumn{1}{c|}{1.232} & \multicolumn{1}{c|}{0.595} & \multicolumn{1}{c|}{0.573} & \multicolumn{1}{c|}{0.355} & \multicolumn{1}{c|}{0.104} & \multicolumn{1}{c|}{0.258} & \multicolumn{1}{c|}{0.091} & \multicolumn{1}{c|}{0.058} & \multicolumn{1}{c}{0.086} \\ 
 
\midrule
R2Gen               & \multicolumn{1}{c|}{0.080} & \multicolumn{1}{c|}{0.315} & \multicolumn{1}{c|}{0.291} & \multicolumn{1}{c|}{0.530} & \multicolumn{1}{c|}{0.463} & \multicolumn{1}{c|}{0.338} & \multicolumn{1}{c|}{1.603} & \multicolumn{1}{c|}{0.601} & \multicolumn{1}{c|}{0.587} & \multicolumn{1}{c|}{0.348} & \multicolumn{1}{c|}{0.039} & \multicolumn{1}{c|}{0.188} & \multicolumn{1}{c|}{0.133} & \multicolumn{1}{c|}{0.152} & \multicolumn{1}{c}{0.178} \\
R2GenCMN  & \multicolumn{1}{c|}{0.080} & \multicolumn{1}{c|}{\textbf{0.321}} & \multicolumn{1}{c|}{\underline{0.363}} & \multicolumn{1}{c|}{\textbf{0.618}} & \multicolumn{1}{c|}{0.493} & \multicolumn{1}{c|}{\textbf{0.385}} & \multicolumn{1}{c|}{1.406} & \multicolumn{1}{c|}{0.625} & \multicolumn{1}{c|}{0.601} & \multicolumn{1}{c|}{0.379} & \multicolumn{1}{c|}{\underline{0.118}} & \multicolumn{1}{c|}{0.287} & \multicolumn{1}{c|}{0.243} & \multicolumn{1}{c|}{\textbf{0.170}} & \multicolumn{1}{c}{\textbf{0.214}} \\
MPO                 & \multicolumn{1}{c|}{0.076} & \multicolumn{1}{c|}{0.299} & \multicolumn{1}{c|}{0.344} & \multicolumn{1}{c|}{0.605} & \multicolumn{1}{c|}{0.475} & \multicolumn{1}{c|}{0.317} & \multicolumn{1}{c|}{1.352} & \multicolumn{1}{c|}{0.589} & \multicolumn{1}{c|}{0.551} & \multicolumn{1}{c|}{0.336} & \multicolumn{1}{c|}{0.067} & \multicolumn{1}{c|}{0.222} & \multicolumn{1}{c|}{0.017} & \multicolumn{1}{c|}{0.013} & \multicolumn{1}{c}{0.017} \\
ESC RL              & \multicolumn{1}{c|}{0.073} & \multicolumn{1}{c|}{0.293} & \multicolumn{1}{c|}{0.333} & \multicolumn{1}{c|}{0.585} & \multicolumn{1}{c|}{0.481} & \multicolumn{1}{c|}{0.295} & \multicolumn{1}{c|}{1.109} & \multicolumn{1}{c|}{0.596} & \multicolumn{1}{c|}{0.557} & \multicolumn{1}{c|}{0.351} & \multicolumn{1}{c|}{0.102} & \multicolumn{1}{c|}{0.272} & \multicolumn{1}{c|}{0.136} & \multicolumn{1}{c|}{0.062} & \multicolumn{1}{c}{0.095} \\
EditGRPO            & \multicolumn{1}{c|}{0.046} & \multicolumn{1}{c|}{0.248} & \multicolumn{1}{c|}{0.310} & \multicolumn{1}{c|}{0.603} & \multicolumn{1}{c|}{0.463} & \multicolumn{1}{c|}{0.334} & \multicolumn{1}{c|}{1.263} & \multicolumn{1}{c|}{0.645} & \multicolumn{1}{c|}{0.606} & \multicolumn{1}{c|}{0.347} & \multicolumn{1}{c|}{0.095} & \multicolumn{1}{c|}{0.190} & \multicolumn{1}{c|}{0.144} & \multicolumn{1}{c|}{0.104} & \multicolumn{1}{c}{0.112} \\
DPO (GREEN)         & \multicolumn{1}{c|}{0.044} & \multicolumn{1}{c|}{0.235} & \multicolumn{1}{c|}{0.300} & \multicolumn{1}{c|}{0.581} & \multicolumn{1}{c|}{0.463} & \multicolumn{1}{c|}{\underline{0.363}} & \multicolumn{1}{c|}{1.385} & \multicolumn{1}{c|}{\textbf{0.676}} & \multicolumn{1}{c|}{\underline{0.633}} & \multicolumn{1}{c|}{0.358} & \multicolumn{1}{c|}{0.067} & \multicolumn{1}{c|}{0.108} & \multicolumn{1}{c|}{0.122} & \multicolumn{1}{c|}{0.107} & \multicolumn{1}{c}{0.097} \\ 
 
\midrule
Qwen3.5-122B        & \multicolumn{1}{c|}{0.028} & \multicolumn{1}{c|}{0.256} & \multicolumn{1}{c|}{0.237} & \multicolumn{1}{c|}{0.524} & \multicolumn{1}{c|}{0.403} & \multicolumn{1}{c|}{0.211} & \multicolumn{1}{c|}{0.881} & \multicolumn{1}{c|}{0.559} & \multicolumn{1}{c|}{0.601} & \multicolumn{1}{c|}{0.376} & \multicolumn{1}{c|}{0.108} & \multicolumn{1}{c|}{0.285} & \multicolumn{1}{c|}{0.198} & \multicolumn{1}{c|}{0.140} & \multicolumn{1}{c}{0.183} \\
InternVL-241B       & \multicolumn{1}{c|}{0.033} & \multicolumn{1}{c|}{0.283} & \multicolumn{1}{c|}{0.244} & \multicolumn{1}{c|}{0.531} & \multicolumn{1}{c|}{0.423} & \multicolumn{1}{c|}{0.211} & \multicolumn{1}{c|}{0.865} & \multicolumn{1}{c|}{0.610} & \multicolumn{1}{c|}{0.619} & \multicolumn{1}{c|}{0.275} & \multicolumn{1}{c|}{0.115} & \multicolumn{1}{c|}{\underline{0.292}} & \multicolumn{1}{c|}{\underline{0.244}} & \multicolumn{1}{c|}{0.124} & \multicolumn{1}{c}{0.208} \\ \midrule
DPO (Our rewards)   & \multicolumn{1}{c|}{0.033} & \multicolumn{1}{c|}{0.243} & \multicolumn{1}{c|}{0.217} & \multicolumn{1}{c|}{0.451} & \multicolumn{1}{c|}{0.478} & \multicolumn{1}{c|}{0.260} & \multicolumn{1}{c|}{\underline{0.787}} & \multicolumn{1}{c|}{0.576} & \multicolumn{1}{c|}{0.351} & \multicolumn{1}{c|}{0.180} & \multicolumn{1}{c|}{0.072} & \multicolumn{1}{c|}{0.116} & \multicolumn{1}{c|}{0.175} & \multicolumn{1}{c|}{0.066} & \multicolumn{1}{c}{0.093} \\
Ours                & \multicolumn{1}{c|}{\underline{0.088}} & \multicolumn{1}{c|}{0.291} & \multicolumn{1}{c|}{0.301} & \multicolumn{1}{c|}{0.583} & \multicolumn{1}{c|}{\underline{0.494}} & \multicolumn{1}{c|}{0.306} & \multicolumn{1}{c|}{1.162} & \multicolumn{1}{c|}{0.619} & \multicolumn{1}{c|}{\textbf{0.648}} & \multicolumn{1}{c|}{\textbf{0.401}} & \multicolumn{1}{c|}{\textbf{0.122}} & \multicolumn{1}{c|}{\textbf{0.310}} & \multicolumn{1}{c|}{\textbf{0.246}} & \multicolumn{1}{c|}{\underline{0.129}} & \multicolumn{1}{c}{\underline{0.212}} \\
\bottomrule[1.5pt]
\end{tabular}%
}
\caption{Report-level IU-Xray results, all methods re-scored under one unified protocol. NLG (left) and clinical-efficacy (right) metrics; RadCliQ ($\downarrow$) is lower-is-better, all others higher-is-better. \textbf{Bold} best, \underline{underline} runner-up per column. \emph{DPO (Our rewards)} is an optimizer-swap ablation under our reward.}
\label{tab:iuxray}
\end{table*}
\section{Experimental Setup}
 \subsection{Datasets and Baselines}
We evaluate HERO on two widely used radiology report generation \textbf{benchmarks}.
\textbf{MIMIC-CXR}~\citep{johnson2024mimic} contains 377,110 chest X-ray images and 227,835 radiology reports. 
We use the ``Findings'' section of each report as the \texttt{<think>} target and derive the \texttt{<answer>} target by applying CheXbert~\citep{smit2020chexbert} to extract 14 structured disease labels, which are formatted as JSON outputs.
\textbf{IU X-Ray}~\citep{demner2015preparing} contains 7,470 images and 3,955 reports. Additional preprocessing details are provided in the Appendix.

We compare HERO with three groups of \textbf{baselines}.
\textbf{Specialized medical MLLMs}: 
R2Gen, R2GenCMN, LLaVA-Rad~\citep{zambrano2025clinically}, MPO~\citep{xiao2025radiology}, ESC-RL, DPO, and EditGRPO~\citep{zhang2025editgrpo}.
\textbf{General-domain MLLMs}: LLaVA and Qwen-2.5-VL-Instruct, finetuned on each dataset using the same multi-task objective as HERO.
\textbf{Inference-only foundation models}: Qwen-3.5-VL (122B)~\citep{bai2025qwen3} and InternVL-3.5 (241B)~\citep{wang2025internvl3}, evaluated in the zero-shot setting as large-scale references.


\label{sec:scs}
\begin{table}[t]
\centering
\resizebox{\columnwidth}{!}{%
\begin{tabular}{l|cccc|cc|ccccc}
\toprule
 & {L15}
 & {LNv}
 & {LNm}
 & {Q25}
 & {Q35}
 & {InVL}
 & {ESC}
 & {MPO}
 & {Edit}
 & {DPO}
 & {\textbf{Ours}} \\
\midrule
\multicolumn{12}{l}{\emph{Answer-Level: Clinical Efficacy}}\\
Micro-F1(14)     & 0.580 & 0.584 & \underline{0.592} & 0.535 & 0.497 & 0.584 & 0.554 & 0.577 & 0.572 & 0.390 & \textbf{0.607} \\
avg-Macro-F1(14) & 0.407 & 0.398 & \textbf{0.420} & 0.363 & 0.347 & 0.364 & 0.382 & 0.361 & 0.367 & 0.275 & \underline{0.412} \\
w-Macro-F1(14)   & 0.527 & \underline{0.565} & 0.521 & 0.549 & 0.488 & 0.545 & 0.557 & 0.544 & 0.542 & 0.381 & \textbf{0.575} \\
Micro-F1(5)      & \underline{0.596} & 0.585 & 0.583 & 0.559 & 0.473 & 0.481 & 0.571 & 0.564 & 0.560 & 0.469 & \textbf{0.618} \\
avg-Macro-F1(5)  & \underline{0.520} & 0.496 & 0.518 & 0.473 & 0.451 & 0.466 & 0.481 & 0.502 & 0.478 & 0.448 & \textbf{0.525} \\
w-Macro-F1(5)    & \textbf{0.597} & 0.575 & 0.579 & 0.550 & 0.488 & 0.519 & 0.560 & 0.556 & 0.548 & 0.481 & \underline{0.589} \\
\midrule
\multicolumn{12}{l}{\emph{SCS: Consistency (think vs answer)}}\\
Micro-F1(14)     & 0.960 & 0.579 & 0.947 & 0.953 & 0.804 & 0.595 & 0.955 & \underline{0.963} & 0.947 & 0.478 & \textbf{0.980} \\
avg-Macro-F1(14) & 0.947 & 0.578 & \underline{0.954} & \textbf{0.961} & 0.542 & 0.559 & 0.952 & 0.943 & 0.949 & 0.385 & 0.845 \\
w-Macro-F1(14)   & 0.959 & 0.644 & 0.947 & 0.953 & 0.765 & 0.666 & 0.955 & \underline{0.963} & 0.948 & 0.486 & \textbf{0.979} \\
Micro-F1(5)      & 0.991 & 0.532 & 0.990 & 0.989 & 0.825 & 0.685 & \underline{0.992} & 0.989 & 0.987 & 0.585 & \textbf{0.995} \\
avg-Macro-F1(5)  & \underline{0.993} & 0.530 & 0.989 & 0.985 & 0.760 & 0.755 & \underline{0.993} & 0.987 & 0.979 & 0.535 & \textbf{0.994} \\
w-Macro-F1(5)    & 0.991 & 0.606 & 0.990 & 0.989 & 0.804 & 0.896 & \underline{0.992} & 0.989 & 0.987 & 0.563 & \textbf{0.995} \\
\midrule
\multicolumn{12}{l}{\emph{Rationale-Supported Score}}\\
$R_6^{\text{gt}}$ & 0.661 & 0.660 & 0.656 & 0.647 & 0.671 & 0.707 & 0.648 & 0.655 & 0.648 & \underline{0.741} & \textbf{0.773} \\
\midrule
\multicolumn{12}{l}{\emph{Diagnosis--Evidence}}\\
SDP $\uparrow$     & 0.580 & 0.282 & 0.555 & 0.574 & 0.384 & 0.479 & \underline{0.583} & 0.544 & 0.547 & 0.286 & \textbf{0.590} \\
SDR $\uparrow$     & 0.478 & 0.232 & \underline{0.481} & 0.415 & 0.382 & 0.428 & 0.433 & 0.418 & 0.427 & 0.323 & \textbf{0.611} \\
UDR $\downarrow$   & \underline{0.002} & 0.499 & \underline{0.002} & 0.010 & 0.211 & 0.047 & 0.006 & 0.013 & 0.003 & 0.400 & \textbf{0.001} \\
FRR $\downarrow$   & \underline{0.515} & 0.544 & 0.533 & 0.516 & 0.581 & 0.683 & 0.523 & 0.529 & 0.561 & 0.610 & \textbf{0.510} \\
\bottomrule
\end{tabular}}
\caption{Answer-level clinical efficacy (answer JSON vs.\ GT, ignore convention), think--answer self-consistency (SCS, GT-free), and evidence metrics on MIMIC-CXR ($N{=}2047$, greedy). SCS and $R_6^{\text{gt}}$ coincide with our reward terms $R_5$ and $R_6$ and are thus optimized by our method. L15 = LLaVA-1.5, LNv/LNm = LLaVA-Next (Vicuna/Mistral), Q25 = Qwen-2.5-VL-7B (SFT on our data), Q35 = Qwen-3.5-122B, InVL = InternVL-241B, ESC = ESC-RL, Edit = EditGRPO, DPO = DPO under our reward (answer JSON parseable for 36\% of studies). \textbf{Bold} best, \underline{underline} runner-up per row.}
\label{tab:scs}
\end{table}

\subsection{Evaluation Metrics}
\label{sec:metrics}
We evaluate HERO across three categories under the standard CheXbert \emph{ignore} convention (excluding uncertain labels; see Appendix).
\textbf{(1) Clinical efficacy:} Micro, macro, and weighted F1 (5- and 14-condition sets) at the answer level (\texttt{<answer>} JSON vs.\ GT) and report level (CheXbert on \texttt{<think>}), plus SRR-BERT weighted F1~\citep{delbrouck2025automated}, RadGraph F1~\citep{delbrouck2022improving}, GREEN~\citep{ostmeier2024green}, RadCliQ-v1~\citep{yu2023evaluating} ($\downarrow$), and RateScore~\citep{zhao2024ratescore}.
\textbf{(2) NLG overlap:} BLEU-4, ROUGE-1/2/L, METEOR, and BERTScore between \texttt{<think>} and the GT report.
\textbf{(3) Reasoning consistency and rationale faithfulness:} We report our GT-free \emph{Self-Consistency Score} (SCS)---the micro, macro, and weighted F1 (5 and 14 conditions) between \texttt{<think>} CheXbert labels and \texttt{<answer>} predictions---and the rationale faithfulness score ($R_6$). Over the 12 disease labels, defining \texttt{<think>} rationale ($\psi=1$) when the labeler marks a disease present or uncertain in \texttt{<think>} (excluding GT-uncertain pairs from SDP/FRR), we also report:
\textbf{SDP} ($\uparrow$) $= \#(A{=}1 \wedge \psi{=}1 \wedge \text{GT}{=}1)/\#(A{=}1)$;
\textbf{SDR} ($\uparrow$) $= \#(A{=}1 \wedge \psi{=}1 \wedge \text{GT}{=}1)/\#(\text{GT}{=}1)$;
\textbf{UDR} ($\downarrow$, GT-free) $= \#(A{=}1 \wedge \psi{=}0)/\#(A{=}1)$; and
\textbf{FRR} ($\downarrow$) $= \#(\psi{=}1 \wedge \text{GT}{\neq}1)/\#(\psi{=}1)$, where $A$ denotes the \texttt{<answer>} prediction.

\section{Results and Analyses}
\subsection{Comparative Analysis on Reports}
Overall, HERO achieves the strongest clinical efficacy on MIMIC-CXR (Table~\ref{tab:mimic}), ranking first on most clinically meaningful metrics while remaining competitive on report-overlap metrics. Specifically, HERO obtains the best RadGraph F1 ($0.354$), RateScore ($0.649$), and four of the six CheXpert F1 variants, demonstrating consistent improvements in diagnostic correctness over existing RRG methods. Although HERO does not always achieve the highest lexical-overlap scores (e.g., BLEU-4 and ROUGE-L), it leads semantics-oriented metrics such as BERTScore and METEOR, suggesting that the proposed optimization primarily improves clinical accuracy rather than surface-level text similarity.

These gains arise from the proposed optimization rather than model scale or reward design. Despite using a 7B backbone, HERO substantially outperforms much larger foundation models (e.g., Qwen3.5-122B and InternVL-241B) on clinical metrics. Moreover, replacing HERO with DPO under the same reward function consistently degrades performance, indicating that the improvements primarily stem from the proposed factorized credit assignment rather than the reward itself. 
Methods optimized for individual objectives (e.g., DPO (GREEN)) remain strongest on their target metrics but do not match HERO in overall clinical efficacy.

The same trend generalizes to IU-Xray (Table~\ref{tab:iuxray}). HERO again achieves the strongest clinical performance while remaining competitive on report-overlap metrics, indicating that the proposed optimization improves diagnostic prediction rather than overfitting to a particular dataset or metric. The optimizer comparison is even more pronounced: under the same reward formulation, DPO degrades substantially whereas HERO consistently maintains superior clinical performance, demonstrating that the proposed factorized policy optimization generalizes better than conventional preference optimization across datasets.

 \begin{table}[t]
\centering
\resizebox{\columnwidth}{!}{%
\begin{tabular}{l|cccc|ccccc}
\toprule[1.5pt]
 & \multicolumn{4}{c|}{Cumulative} & \multicolumn{5}{c}{Ablation} \\
\cmidrule(lr){2-5}\cmidrule(lr){6-10}
Metric & Base & +L1 & +L2 & Full & $-$L1 & $-$L2 & $-$Gate & L3o & L2o \\
\midrule
\multicolumn{10}{l}{\emph{NLG Metrics}}\\
B-4        & 0.106 & \underline{0.111} & 0.107 & 0.102 & \textbf{0.119} & 0.103 & 0.097 & 0.105 & 0.105 \\
METEOR     & 0.329 & \underline{0.353} & 0.342 & 0.334 & \textbf{0.356} & 0.343 & 0.329 & 0.328 & 0.327 \\
ROUGE-L    & \underline{0.357} & 0.347 & 0.335 & 0.340 & \textbf{0.369} & 0.338 & 0.327 & 0.356 & 0.355 \\
BERTScore  & 0.610 & 0.606 & 0.602 & \underline{0.617} & \textbf{0.618} & 0.589 & 0.579 & 0.612 & 0.607 \\
SRR-bert   & 0.479 & 0.472 & 0.465 & \textbf{0.487} & 0.462 & 0.466 & 0.453 & \underline{0.486} & 0.483 \\
\midrule
\multicolumn{10}{l}{\emph{Clinical Efficacy}}\\
RadGraph            & 0.317 & 0.348 & 0.343 & \textbf{0.354} & \underline{0.352} & 0.311 & 0.298 & 0.322 & 0.313 \\
RadCliQ $\downarrow$ & 1.342 & 1.372 & 1.345 & 1.178 & 1.403 & \underline{1.169} & \textbf{1.036} & 1.362 & 1.312 \\
RateScore           & 0.635 & 0.644 & \textbf{0.652} & \underline{0.649} & 0.642 & 0.602 & 0.600 & 0.638 & 0.632 \\
GREEN               & 0.483 & 0.501 & \textbf{0.518} & 0.498 & 0.497 & \underline{0.510} & 0.461 & 0.482 & 0.484 \\
Micro-F1(14)        & 0.558 & 0.572 & 0.569 & \textbf{0.574} & 0.565 & \underline{0.573} & 0.561 & 0.562 & 0.554 \\
avg-Macro-F1(14)    & 0.395 & 0.383 & 0.386 & \textbf{0.413} & 0.376 & 0.399 & 0.382 & \underline{0.409} & 0.405 \\
w-Macro-F1(14)      & 0.523 & 0.520 & 0.526 & \underline{0.530} & 0.518 & 0.528 & 0.510 & \textbf{0.533} & 0.523 \\
Micro-F1(5)         & 0.614 & 0.614 & 0.606 & \textbf{0.618} & 0.601 & \underline{0.616} & 0.602 & 0.604 & 0.606 \\
avg-Macro-F1(5)     & 0.521 & 0.522 & 0.504 & \underline{0.525} & 0.498 & 0.524 & 0.521 & 0.523 & \textbf{0.529} \\
w-Macro-F1(5)       & \underline{0.597} & \underline{0.597} & 0.587 & 0.589 & 0.582 & \textbf{0.598} & 0.589 & 0.591 & 0.594 \\
\midrule
\multicolumn{10}{l}{\emph{Rationale-Supported Score}}\\
$R_6^{\text{gt}}$ & 0.690 & 0.718 & 0.724 & \textbf{0.773} & 0.720 & \underline{0.755} & 0.734 & 0.686 & 0.695 \\
\bottomrule[1.5pt]
\end{tabular}%
}
\caption{Ablation of the three-layer framework on MIMIC-CXR (transposed; metrics in rows, variants in columns). \textbf{Bold} best, \underline{underline} runner-up per metric. \emph{Cumulative}: Base $\rightarrow$ +L1 $\rightarrow$ +L2 $\rightarrow$ Full ($=$+L3). \emph{Ablations}: $-$L1/$-$L2/$-$Gate are leave-one-out from Full; L3o/L2o are single-layer variants.}
\label{tab:ablation}
\end{table}
\subsection{Reasoning Consistency and Rationale Faithfulness}
\label{sec:consistency}

Table~\ref{tab:scs} evaluates reasoning quality across four dimensions: answer-level efficacy, SCS, rationale-supported score ($R_6^{\text{gt}}$), and independent CheXpert diagnosis--rationale metrics. HERO leads answer-level diagnosis, ranking first on four of six F1 metrics---including Micro-F1(14) ($0.607$) and Micro-F1(5) ($0.618$)---and runner-up on the remaining two.

Crucially, Table~\ref{tab:scs} shows that self-consistency alone does not guarantee faithful reasoning. Most competitive baselines already achieve near-saturated SCS ($>0.94$ on Micro-F1(14)) and low unsupported diagnosis rates ($\text{UDR} \le 0.013$), yet remain low on $R_6^{\text{gt}}$ ($0.647$--$0.661$ vs.\ our $0.773$) and exhibit high fabricated rationale rates ($\text{FRR}\ 0.515$--$0.561$). ESC-RL exemplifies this gap: despite reaching $0.955$ SCS and $0.583$ SDP, it scores only $0.648$ on $R_6^{\text{gt}}$ and $0.433$ on SDR. Conversely, InternVL-241B and DPO (Our rewards) attain higher $R_6^{\text{gt}}$ ($0.707$ and $0.741$) but suffer from collapsed consistency (SCS $0.595$ and $0.478$), high UDR ($0.047$ and $0.400$), and severe rationale fabrication ($\text{FRR}\ 0.683$ and $0.610$). 

In contrast, HERO is the only method that excels across all four blocks: it achieves the highest Micro-F1(14) SCS ($0.980$) and $R_6^{\text{gt}}$ ($0.773$), while ranking best on all four independent diagnosis--rationale metrics. Notably, HERO boosts Supported Diagnosis Recall (SDR) to $0.611$ ($+0.130$ over the runner-up LLaVA-Next-Mistral at $0.481$) while simultaneously achieving the highest SDP ($0.590$) and the lowest UDR ($0.001$) and FRR ($0.510$), confirming that Factor-3 gating genuinely grounds diagnoses in valid visual findings rather than merely aligning unverified narratives.

\subsection{Ablation of the Three Factors}
\label{sec:ablation}
Setting each HERO factor to its neutral value enables exact ablations (Table~\ref{tab:ablation}; Appendix). Cumulatively adding factors to the GRPO baseline shows complementary roles: Factor~1 (segment advantage) drives initial clinical gains in RadGraph ($0.317\!\rightarrow\!0.348$), Micro-F1(14) ($0.558\!\rightarrow\!0.572$), and GREEN ($0.483\!\rightarrow\!0.501$); Factor~2 (token reweighting) lifts semantic/entity scores (RateScore $0.644\!\rightarrow\!0.652$, GREEN $0.501\!\rightarrow\!0.518$) with flat F1; and Factor~3 (rationale-supported gate) yields the largest jump in avg-Macro-F1(14) ($0.386\!\rightarrow\!0.413$), Micro-F1(5) ($0.606\!\rightarrow\!0.618$), and RadGraph ($0.343\!\rightarrow\!0.354$), for total gains of $+0.037$ RadGraph and $\sim\!+0.017$ Micro-/avg-Macro-F1(14).

In leave-one-out ablations, removing Factor~1 trades clinical accuracy (avg-Macro-F1(14) $0.413\!\rightarrow\!0.376$) for lexical overlap (BLEU-4 $0.102\!\rightarrow\!0.119$, ROUGE-L $0.340\!\rightarrow\!0.369$), while removing Factor~2 sharply drops RadGraph ($0.354\!\rightarrow\!0.311$). Removing the gate (retaining $R_6$) causes the worst degradation---RadGraph ($0.354\!\rightarrow\!0.298$), Micro-F1(14) ($0.574\!\rightarrow\!0.561$), and GREEN ($0.498\!\rightarrow\!0.461$; composite RadCliQ drops $1.178\!\rightarrow\!1.036$)---proving the gating \emph{mechanism} is essential beyond $R_6$ itself. Rationale faithfulness ($R_6^{\text{gt}}$) likewise rises monotonically ($0.690\!\rightarrow\!0.718\!\rightarrow\!0.724\!\rightarrow\!0.773$), drops most without the gate ($0.773\!\rightarrow\!0.734$), and stays near baseline under single-factor L2o/L3o ($0.695$/$0.686$), confirming that strong rationale faithfulness and clinical efficacy require all three factors working together.
\section{Conclusion}
\label{sec:conclusion}
 

We presented HERO, a hierarchical GRPO framework that treats supervision in reason-then-summarize radiology report generation as heterogeneous rather than collapsing it into a single broadcast advantage. 
HERO factorizes policy optimization through three complementary factors: segment-level optimization for reasoning and diagnosis, token-level optimization for diagnostic salience, and a rationale-supported completion factor that reinforces diagnoses supported by generated reasoning. 
This design effectively mitigates clinical hallucinations by discouraging unsupported rationales. 
Experiments on MIMIC-CXR and IU-Xray demonstrate state-of-the-art clinical efficacy, with HERO emerging as the only system that is simultaneously think--answer consistent and rationale-faithful. 
Future work will further isolate the contribution of rationale-supported optimization through consistency-only ablations and extend the proposed factorized policy optimization framework to broader structured clinical generation tasks.

\section*{Limitations}

While HRRO demonstrates strong clinical efficacy and rationale faithfulness, several limitations remain. 
First, our reward formulation heavily relies on automated natural language processing tools, specifically CheXbert, for extracting structured labels from both ground-truth reports and generated rationales. Although CheXbert is highly accurate, it is not flawless; any inherent extraction errors or biases in the labeler act as a ceiling on our reward quality and could be inadvertently amplified during reinforcement learning. 
Second, our current experimental scope is limited to single-image chest X-rays. In clinical practice, radiologists routinely compare current scans with a patient's prior imaging and historical health records. Because our data preprocessing explicitly removes longitudinal comparisons to frame the task as an image--text mapping, the model lacks the temporal context necessary for comprehensive real-world reporting. 
Third, our method currently lacks spatial visual grounding. While HRRO ensures that diagnostic conclusions are faithfully supported by the generated textual rationale, it does not explicitly ground these textual findings to specific regions within the chest X-ray (e.g., via bounding boxes or heatmaps). We leave the integration of dense visual evidence grounding—aligning the reasoning process directly with localized anatomical features—as an important direction for future work.
Fourth, extending HRRO to volumetric modalities (e.g., CT or MRI) or much longer document generation remains unexplored. 
Finally, while HRRO is highly memory-efficient compared to critic-based PPO, group-relative policy optimization still requires sampling multiple completions per image during training, presenting a higher computational and memory overhead than standard supervised fine-tuning.

\section*{Ethic Statements}

The development of Multimodal Large Language Models for medical applications carries significant ethical implications. 
First and foremost, the model presented in this work is intended purely as an assistive research tool to explore reasoning-aligned optimization. It is not designed, nor is it ready, for autonomous clinical deployment. Despite our method substantially reducing rationale--diagnosis decoupling and clinical hallucinations, Multimodal LLMs remain susceptible to occasional factual errors or omissions. Any generated radiology report must be rigorously reviewed and verified by qualified medical professionals to prevent potential patient harm.
Furthermore, our models are trained and evaluated on specific, geographically localized datasets (e.g., MIMIC-CXR and IU-Xray). Such datasets inherently reflect the demographic distributions, clinical workflows, and stylistic reporting preferences of their respective institutions. Consequently, the model may exhibit demographic biases or unequal performance across underrepresented patient populations. Future translational research must thoroughly audit these models for algorithmic fairness and robustness across diverse, multi-institutional cohorts before they can be considered for real-world clinical integration. All datasets used in this study are publicly available and pre-deidentified; thus, no new privacy or patient consent issues are introduced.


\bibliography{custom, references}

\appendix

\section{Implementation Details}
\label{sec:appendix-method}

\subsection{Rationale-Faithful Scoring and Gate Parameters}
For each pathology \(k\), let \(a_k\) denote the answer label, \(g_k\) denote the ground-truth label, and \(\psi_k\) denote the \textit{think} mention indicator. Scoring matrix is Table ~\ref{tab:pathology_scoring}.

\begin{table*}[t] 
\centering  
\small
\begin{tabular}{ccccp{6.2cm}} 
\toprule Answer \(a_k\) & Think mention \(\psi_k\) & GT \(g_k\) & Score \(s_k\) & Interpretation \\ \midrule 1 & 1 & 1 & \(+1.0\) & True positive correctly identified and explained. \\ 1 & 0 & 1 & \(+0.5\) & True positive, but the supporting narrative is missing. \\ 1 & 1 & 0 & \(-0.5\) & False positive accompanied by a fabricated explanation. \\ 1 & 0 & 0 & \(-1.0\) & Pure false positive without a supporting narrative. \\ 0 & 0 & 0 & \(+1.0\) & True negative correctly left unmentioned. \\ 0 & 1 & 0 & \(-1.0\) & Mentioned in the reasoning but not claimed in the answer. \\ \(0\) or \texttt{null} & 0 & 1 & \(-1.0\) & False negative: the positive pathology is missed entirely. \\ \(0\) or \texttt{null} & 1 & 1 & \(-0.5\) & The pathology is recognized in the reasoning but omitted from the answer. \\ \bottomrule 
\end{tabular} 
\caption{Scoring Matrix: Pathology-level scoring rules.}
\label{tab:pathology_scoring} 
\end{table*}

Based on this scoring matrix, the two constants in Eq. 11 are fixed a priori from the range of the faithfulness score rather than tuned against validation performance.
Under the rationale-faithful scoring matrix each per-label score satisfies
$s_k\in\{-1,-\tfrac12,+\tfrac12,+1\}$, so the composite score is
confined to the symmetric interval $C(y_i)\in[-1,1]$, with $C=0$ the
indifference point at which a completion's supported and unsupported findings
exactly offset. Two requirements then determine $\delta$ and $\gamma$.

\emph{Threshold.} The gate should neither reward nor suppress a completion at
the indifference point, i.e.\ $\phi_{\mathrm{rat}}=\tfrac12$ when $C=0$. Since
$\sigma(0)=\tfrac12$, this fixes $\delta=0$ exactly; no other value is
consistent with the symmetry of the score.

\emph{Sharpness.} The gate should traverse most of its codomain over the
attainable score range while remaining smooth. Writing the $10\%$--$90\%$
transition band of $\sigma$ as $|C-\delta|\le\ln 9/\gamma$, the choice
$\gamma=5$ places that band at $\pm0.44$. Two properties follow. First, the
gate is not saturated: it spans $\phi\in[0.007,0.993]$, i.e.\ $98.6\%$ of
$(0,1)$, over $C\in[-1,1]$ (Table~\ref{tab:gate}), so the full range of rationale
faithfulness is expressed rather than collapsed onto a hard mask. Second, the gate
is not brittle: with all $14$ labels evaluated, flipping one label between
fully supported and fully unsupported moves $C$ by $2/|\mathcal{P}'|\approx0.14$,
so the transition band spans roughly six label flips. The gate therefore
discriminates at the granularity of the overall rationale, not of a
single keyword match---a deliberate choice, since $\psi_k$ is a lexical
detector and individual mentions are recoverable errors. Larger $\gamma$ would
approach a step function and reintroduce the discrete masking artifacts that
motivated a sigmoid in the first place; smaller $\gamma$ would compress $\phi$
around $\tfrac12$ and render the gate inert.
\begin{table*}[t]
\centering
\begin{tabular}{lccccccc}
\toprule
$C(y_i)$ & $-1.0$ & $-0.5$ & $-0.25$ & $0.0$ & $0.25$ & $0.5$ & $1.0$ \\
\midrule
$\phi_{\text{evidence}}$ & $0.007$ & $0.076$ & $0.223$ & $0.500$ & $0.777$ & $0.924$ & $0.993$ \\
\bottomrule
\end{tabular}
\caption{Gate response $\phi_{\text{evidence}}=\sigma(\gamma(C-\delta))$ at
$\gamma{=}5.0$, $\delta{=}0.0$. The gate is smooth and unsaturated across the
attainable score range $C\in[-1,1]$, suppressing completions whose positive
findings lack support in think without hard-masking them.}
\label{tab:gate}
\end{table*}

\subsubsection{Why the gate need not be tuned.}
Two structural properties make method insensitive by construction to
the precise value of $\gamma$. (i)~The evidence score enters the objective
\emph{twice}: additively through the reward component $R_6$, which shapes the
group-normalized advantage independently of $\gamma$, and multiplicatively
through Eq.2. The gate modulates the strength of a signal that
is already present, so a mis-specified $\gamma$ attenuates rather than removes
the evidence supervision. (ii)~The parameterization admits an exact identity
setting (Proposition~3): $\gamma\to0^{+}$ recovers the ungated objective
continuously. The configuration space is therefore not knife-edge---there is a
smooth path from the operating point to the ablated model, with no
discontinuity in between. We emphasize that this is an argument about the
\emph{form} of the gate, not a measured robustness result; the full
hyperparameter disclosure is given in Section~\ref{sec:hparams}.

\subsection{Hyperparameter Sensitivity}
\label{sec:hparams}

\paragraph{Hyperparameters}
The reward weights are
$\bm{w}=(0.05,0.05,0.05,0.12,0.02,0.07,0.10,0.06,0.50)$ for the nine reward
components. The spillover coefficient is $\lambda=0.5$; the answer floor is
$c_{\text{an}}=0.3$; the standard-deviation clamp is
$\epsilon_{\text{std}}=10^{-4}$; the TF--IDF blending coefficient is
$\alpha=0.3$; the pathology-keyword boost is $b=1.5$; and the rationale-gate
sigmoid uses slope $\gamma=5.0$ and threshold $\delta=0.0$.
HERO is built upon Qwen-2.5-VL-7B and trained using a two-stage optimization strategy. We first perform supervised fine-tuning for 2 epochs using AdamW with a learning rate of $2\times10^{-5}$. The model is then optimized with the proposed GRPO objective and composite reward function. Unless otherwise specified, we use a group size of $G=8$, a clipping parameter of $\varepsilon=0.2$, and a KL regularization coefficient of $\beta=0.03$. All experiments are conducted on NVIDIA A100 GPUs with 80GB memory. Additional implementation details and data preprocessing procedures are provided in the Supplementary Material.
\paragraph{Spillover and group variance}
The spillover parameter $\lambda$ resolves an issue of variance rather than scale. Reasoning-only signals exhibit minimal variation across a group of completions for a given image. Empirically, the range of the think composite is an order of magnitude smaller than that of the answer composite ($r^{\text{th}}_{\max}\!\approx\!0.21$ vs.\ $r^{\text{an}}_{\max}\!\approx\!1.14$). Consequently, its group standard deviation approaches zero, causing the clamp in Eq.~(3) to dominate and rendering the standardized think advantage uninformative. By routing the shared components into the think composite with a weight of $\lambda{=}0.5$, we restore a non-degenerate intra-group variance.

\paragraph{Absence of Hyperparameter Overfitting.} 
The performance gains of our method do not stem from extensive hyperparameter search. Every introduced constant---including the gate parameters $\gamma$ and $\delta$, the token-layer constants $\alpha$ and $b$, the segment spillover coefficient, and the auxiliary reward thresholds---was determined based on scale arguments regarding the quantities they control. These values were held constant across the entire ablation study and transferred \emph{unchanged} from MIMIC-CXR to IU-Xray, a dataset with distinctly different label statistics, report lengths, and scales. Consequently, no selection pressure from either test set influenced these parameters. In this precise sense, our method remains effective without per-dataset tuning, ensuring that our ablation ladder reflects the true contribution of each \emph{mechanism} rather than the artifact of a successful parameter search.

\paragraph{Scope of Robustness Claims.} 
Importantly, our findings do not imply that performance is strictly invariant to these constants. We did not systematically train or evaluate alternative hyperparameter values, and we make no explicit claims regarding robustness. It is entirely possible that certain configurations could outperform our defaults, and parameters such as $\alpha$ or $\gamma$ may possess pronounced optima that we have yet to discover. Therefore, our reported results should be interpreted as a \emph{lower bound} of performance achievable without task-specific tuning, rather than the peak capability of an optimized configuration.

\section{Dataset Preprocessing}
To generate concise and coherent medical summaries from unstructured reports, we applied the same preprocessing pipeline to both datasets. First, we removed incomplete reports that lacked either a findings or an impression section, excluded reports with findings sections containing fewer than 10 words, and discarded reports with impression sections containing fewer than 2 words. We then used GPT-4o to further improve summary quality by removing sentences that compared the current examination with the patient’s prior medical history, eliminating de-identification placeholders such as ``\_\_'' while preserving the original meaning, and excluding view-related information because our training is based on image–text pairs. Finally, the cleaned findings and impression sections were combined to produce concise, coherent, and high-quality summaries for each image.

\section{Composability: Exact Ablation Ladder}
\label{app:proofs}

Each factor of HERO recovers a well-defined special case when set to its neutral value,
so ablations are exact rather than approximate. Throughout, the per-token objective is
\begin{equation}
\begin{aligned}
\mathcal{L}(\theta)
=
\frac{1}{N_{\mathrm{tok}}}
\sum_{i,t}
\Big[
&\phi_{\mathrm{rat}}(y_i)\,
s_{i,t}^{\mathrm{IS}}\,
\operatorname{surr}
\big(\rho_{i,t},A_{i,t}\big)
\\
&+\beta\,\mathrm{KL}_{i,t}
\Big]
M_{i,t}^{\mathrm{comp}},
\end{aligned}
\label{eq:compact}
\end{equation}
with $A_{i,t}=A_{i,t}^{\text{seg}}\,\hat w_{i,t}$ and
$A_{i,t}^{\text{seg}}=A_i^{\text{th}}M_{i,t}^{\text{th}}+A_i^{\text{an}}M_{i,t}^{\text{an}}$. Let
$\mathcal{L}_{\text{GRPO}}$ be the standard token-level GRPO/DAPO objective: the same
expression with $A_{i,t}^{\text{seg}}{=}A_i$ (a single scalar group advantage),
$\hat w_t{\equiv}1$, and $\phi_{\mathrm{rat}}{\equiv}1$.

\begin{proposition}[Segment layer]
\label{prop:seg}
If the segment decomposition is disabled ($A_i^{\text{th}}{=}A_i^{\text{an}}{=}A_i$, so
$A_{i,t}^{\text{seg}}{=}A_i$), then $A_{i,t}{=}A_i\hat w_t$ and $\mathcal{L}(\theta)$
reduces to token-reweighted scalar-advantage GRPO; with additionally $\hat w_t{\equiv}1$
it equals $\mathcal{L}_{\text{GRPO}}$.
\end{proposition}
\begin{proof}
A completion token lies in exactly one segment, where the corresponding mask is $1$
(gap tokens carry zero advantage), so $A_{i,t}^{\text{seg}}{=}A_i(M_t^{\text{th}}{+}M_t^{\text{an}}){=}A_i$
on masked tokens, giving $A_{i,t}{=}A_i\hat w_t$. Substitution yields the reweighted
scalar-advantage objective; with $\hat w_t{\equiv}1$, $A_{i,t}{=}A_i$ and the expression
is identical to $\mathcal{L}_{\text{GRPO}}$.
\end{proof}

\begin{proposition}[Token layer]
\label{prop:tok}
If the token weights are uniform ($\hat w_t{\equiv}1$, attained at $\alpha{=}0$,
$b{=}1$), then $A_{i,t}{=}A_{i,t}^{\text{seg}}$ and $\mathcal{L}(\theta)$ equals HERO
without the token layer. Moreover the mean-one normalization guarantees
$\sum_t\hat w_tM_t^{\text{comp}}{=}\sum_tM_t^{\text{comp}}$.
\end{proposition}
\begin{proof}
At $\alpha{=}0,b{=}1$, $w_t^{\text{tok}}{=}1$ for all $t$, so
$\hat w_t{=}\frac{\sum_{t'}M_{t'}^{\text{comp}}}{\sum_{t'}M_{t'}^{\text{comp}}}{=}1$ and
$A_{i,t}{=}A_{i,t}^{\text{seg}}$. For general weights, multiplying the normalization by
$M_t^{\text{comp}}$ and summing over $t$ gives
$\sum_t\hat w_tM_t^{\text{comp}}{=}\sum_tM_t^{\text{comp}}$, so the mean masked weight is
$1$ and the per-completion gradient scale is preserved.
\end{proof}

\begin{proposition}[Rationale-Supported gate]
\label{prop:gate}
If the gate is disabled ($\phi_{\mathrm{rat}}{\equiv}1$), the per-token loss reduces to
$L_{i,t}{=}s_{i,t}^{\mathrm{IS}}\operatorname{surr}(\rho_{i,t},A_{i,t})+\beta\,\mathrm{KL}_{i,t}$,
i.e.\ HERO without the gate; the KL penalty to $\pi_{\text{ref}}$ is unchanged.
\end{proposition}
\begin{proof}
Setting $\phi_{\mathrm{rat}}(y_i){=}1$ removes the leading factor of the policy-gradient
term. As $\phi_{\mathrm{rat}}$ never multiplies $\mathrm{KL}_{i,t}$, the regularization term
is identical with and without the gate.
\end{proof}

\noindent Composing the three, the neutral assignment $A_{i,t}^{\text{seg}}{=}A_i$,
$\hat w_t{\equiv}1$, $\phi_{\mathrm{rat}}{\equiv}1$, $s_{i,t}^{\text{IS}}{\equiv}1$ reduces
$\mathcal{L}(\theta)$ to $\mathcal{L}_{\text{GRPO}}$ exactly. Since the three factors act
on disjoint quantities---a per-segment advantage, a per-token weight, and a
per-completion scalar---they may be toggled in any combination, so every row of the
ablation in the main text corresponds to an exact special case of the objective.

\textbf{D Human Annotation}

To verify that our performance gains are not merely artifacts of the CheXbert labeler, Table \ref{tab:human} presents an evaluation on a human-annotated test set sourced from PhysioNet. We analyze two distinct representations derived from the same model generation against human ground truth: CheXbert labels extracted from the free-text report (“report”) and labels parsed from the structured \texttt{<answer>} JSON (“answer”). This analysis yields two primary conclusions. 

First, our approach maintains a strong performance advantage under human evaluation. It achieves the best overall micro-averaged F1 scores (Micro-F1(14): 0.527; Micro-F1(5): 0.560) and w-Macro-F1(5) (0.567), confirming that our gains are robust and not mere CheXbert-versus-CheXbert artifacts. 

Second, our method demonstrates exceptional internal consistency, which serves as a key diagnostic of its operational goal. While baseline models—especially general-purpose VLMs like Llava-Next(vicuna) and InternVL-241B—exhibit a pronounced performance gap between their textual reports and structured answers (e.g., Llava-Next(vicuna) shows a discrepancy of 0.088 in Micro-F1(14); InternVL-241B shows a gap of 0.057), our model's report and answer views are nearly identical ($|\Delta \text{Micro-F1(14)}| = 0.004$). Given that both representations are decoded from the exact same output, the baseline gaps reveal a fundamental failure to accurately summarize free-text reasoning into structured formats. In contrast, our near-zero discrepancy highlights the effectiveness of the segment-decomposed, rationale-gated objective in harmonizing the model's rationale, structured decisions, and report prose.

\begin{table}[htbp]
\centering
\resizebox{1.0\columnwidth}{!}{%
\begin{tabular}{l|cccccc}
\toprule[1.5pt]
                    & \multicolumn{6}{c}{Clinical Efficacy} \\ \midrule
                    & \multicolumn{1}{c|}{Micro-F1(14)} & \multicolumn{1}{c|}{avg-Macro-F1(14)} & \multicolumn{1}{c|}{w-Macro-F1(14)} & \multicolumn{1}{c|}{Micro-F1(5)} & \multicolumn{1}{c|}{avg-Macro-F1(5)} & \multicolumn{1}{c}{w-Macro-F1(5)} \\ \midrule
Llava-1.5 (report)           & \multicolumn{1}{c|}{0.476} & \multicolumn{1}{c|}{0.354} & \multicolumn{1}{c|}{0.462} & \multicolumn{1}{c|}{0.548} & \multicolumn{1}{c|}{0.481} & \multicolumn{1}{c}{0.557} \\
Llava-1.5 (answer)           & \multicolumn{1}{c|}{0.470} & \multicolumn{1}{c|}{0.341} & \multicolumn{1}{c|}{0.456} & \multicolumn{1}{c|}{0.541} & \multicolumn{1}{c|}{0.473} & \multicolumn{1}{c}{0.552} \\
Llava-Next(vicuna) (report)  & \multicolumn{1}{c|}{0.362} & \multicolumn{1}{c|}{0.299} & \multicolumn{1}{c|}{0.378} & \multicolumn{1}{c|}{0.391} & \multicolumn{1}{c|}{0.389} & \multicolumn{1}{c}{0.443} \\
Llava-Next(vicuna) (answer)  & \multicolumn{1}{c|}{0.450} & \multicolumn{1}{c|}{0.332} & \multicolumn{1}{c|}{0.441} & \multicolumn{1}{c|}{0.518} & \multicolumn{1}{c|}{0.475} & \multicolumn{1}{c}{0.541} \\
Llava-Next(mistral) (report) & \multicolumn{1}{c|}{0.499} & \multicolumn{1}{c|}{0.384} & \multicolumn{1}{c|}{0.484} & \multicolumn{1}{c|}{0.557} & \multicolumn{1}{c|}{0.456} & \multicolumn{1}{c}{0.556} \\
Llava-Next(mistral) (answer) & \multicolumn{1}{c|}{0.483} & \multicolumn{1}{c|}{0.368} & \multicolumn{1}{c|}{0.473} & \multicolumn{1}{c|}{0.548} & \multicolumn{1}{c|}{0.464} & \multicolumn{1}{c}{0.553} \\
Llava-Rad                    & \multicolumn{1}{c|}{0.510} & \multicolumn{1}{c|}{0.381} & \multicolumn{1}{c|}{0.494} & \multicolumn{1}{c|}{0.551} & \multicolumn{1}{c|}{0.460} & \multicolumn{1}{c}{0.553} \\
\midrule
R2Gen                        & \multicolumn{1}{c|}{0.427} & \multicolumn{1}{c|}{0.261} & \multicolumn{1}{c|}{0.385} & \multicolumn{1}{c|}{0.533} & \multicolumn{1}{c|}{0.438} & \multicolumn{1}{c}{0.531} \\
R2GenCMN                     & \multicolumn{1}{c|}{0.425} & \multicolumn{1}{c|}{0.250} & \multicolumn{1}{c|}{0.381} & \multicolumn{1}{c|}{0.531} & \multicolumn{1}{c|}{0.427} & \multicolumn{1}{c}{0.523} \\
MPO (report)                 & \multicolumn{1}{c|}{0.453} & \multicolumn{1}{c|}{0.301} & \multicolumn{1}{c|}{0.420} & \multicolumn{1}{c|}{0.539} & \multicolumn{1}{c|}{0.479} & \multicolumn{1}{c}{0.544} \\
ESC RL (report)              & \multicolumn{1}{c|}{0.466} & \multicolumn{1}{c|}{0.293} & \multicolumn{1}{c|}{0.437} & \multicolumn{1}{c|}{0.554} & \multicolumn{1}{c|}{0.455} & \multicolumn{1}{c}{0.558} \\
EditGRPO (report)            & \multicolumn{1}{c|}{0.455} & \multicolumn{1}{c|}{0.286} & \multicolumn{1}{c|}{0.423} & \multicolumn{1}{c|}{0.547} & \multicolumn{1}{c|}{0.439} & \multicolumn{1}{c}{0.537} \\
\midrule
Qwen3.5-122B (report)        & \multicolumn{1}{c|}{0.349} & \multicolumn{1}{c|}{0.249} & \multicolumn{1}{c|}{0.346} & \multicolumn{1}{c|}{0.450} & \multicolumn{1}{c|}{0.353} & \multicolumn{1}{c}{0.449} \\
Qwen3.5-122B (answer)        & \multicolumn{1}{c|}{0.334} & \multicolumn{1}{c|}{0.275} & \multicolumn{1}{c|}{0.372} & \multicolumn{1}{c|}{0.428} & \multicolumn{1}{c|}{0.414} & \multicolumn{1}{c}{0.477} \\
InternVL-241B (report)       & \multicolumn{1}{c|}{0.349} & \multicolumn{1}{c|}{0.238} & \multicolumn{1}{c|}{0.359} & \multicolumn{1}{c|}{0.476} & \multicolumn{1}{c|}{0.433} & \multicolumn{1}{c}{0.520} \\
InternVL-241B (answer)       & \multicolumn{1}{c|}{0.406} & \multicolumn{1}{c|}{0.296} & \multicolumn{1}{c|}{0.424} & \multicolumn{1}{c|}{0.464} & \multicolumn{1}{c|}{0.455} & \multicolumn{1}{c}{0.527} \\
\midrule
Ours (report)                & \multicolumn{1}{c|}{0.523} & \multicolumn{1}{c|}{0.367} & \multicolumn{1}{c|}{0.499} & \multicolumn{1}{c|}{0.556} & \multicolumn{1}{c|}{0.479} & \multicolumn{1}{c}{0.562} \\
Ours (answer)                & \multicolumn{1}{c|}{0.527} & \multicolumn{1}{c|}{0.368} & \multicolumn{1}{c|}{0.483} & \multicolumn{1}{c|}{0.560} & \multicolumn{1}{c|}{0.483} & \multicolumn{1}{c}{0.567} \\
\bottomrule[1.5pt]
\end{tabular}%
}
\caption{Clinical efficacy on the human-annotation test set ($n{=}500$, greedy). ``(report)'' = CheXbert labels of the generated report (the \texttt{<think>} section; the full report for Llava-Rad, R2Gen and R2GenCMN) vs GT (ignore-uncertain); ``(answer)'' = parsed \texttt{<answer>}-JSON labels vs GT (label\_f1, ignore-uncertain).}
\label{tab:human}
\end{table}

\section{Qualitative Analysis}
\begin{figure*}[t]
\centering
\includegraphics[width=0.8\textwidth]{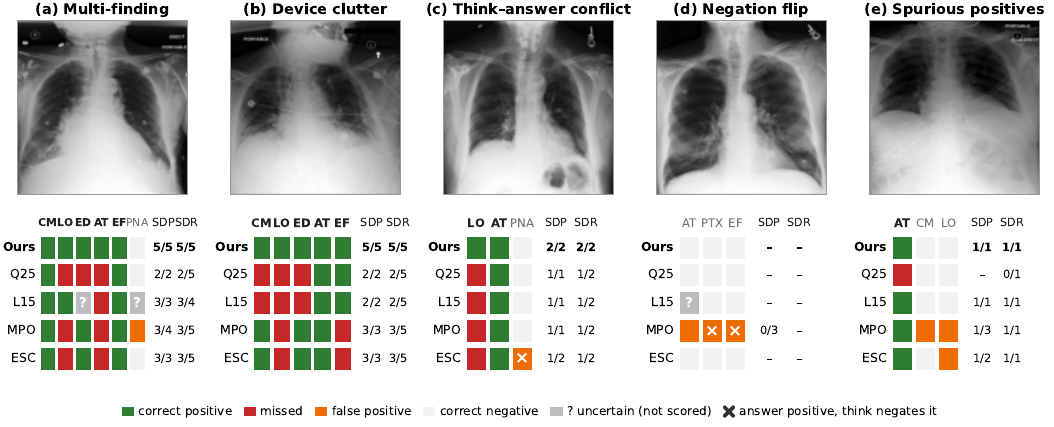}
\caption{Qualitative comparison on MIMIC-CXR. For each study, rows are models and columns are the disease labels that are positive in the ground truth (bold) or predicted by any model. Cell colour encodes the \texttt{<answer>} label against the ground truth (green: correct positive, red: missed, orange: false positive, grey ``?'': uncertain answer, not scored); $\times$ marks a positive answer whose own \texttt{<think>} negates the finding. SDP / SDR = correct diagnoses supported by the think divided by the model's positive diagnoses / by the ground-truth disease positives (-- : empty denominator); support is judged by the CheXpert rule labeler with its negation errors on these studies corrected by manual reading. (a,b) Ours recovers every co-occurring finding; (c,d) baselines contradict their own reasoning; (e) baselines add spurious positives. Labels: CM = Cardiomegaly, LO = Lung Opacity, ED = Edema, AT = Atelectasis, EF = Pleural Effusion, PNA = Pneumonia, PTX = Pneumothorax. Q25 = Qwen-2.5-VL-7B (SFT), L15 = LLaVA-1.5.}
\label{fig:case_study}
\end{figure*}

Figure~\ref{fig:case_study} illustrates three behaviours behind Table~\ref{tab:scs}.
\textbf{Complete diagnosis.} In (a) and (b), each with five co-occurring abnormalities, our model reports all of them
with supporting reasoning (SDR $5/5$), while baselines recover two or three; they negate edema (SFT in (a)) or, in the
ICU film (b), focus on devices and drop parenchymal findings.
\textbf{Reasoning--answer conflicts.} ESC-RL reasons ``without focal consolidation to suggest pneumonia'' yet labels
pneumonia (c), and MPO states ``no pleural effusion or pneumothorax'' yet labels both (d). Our model avoids such
conflicts (test-set UDR $0.001$), the failure mode targeted by the rationale-supported gate.
\textbf{Spurious positives.} In (e), MPO and ESC-RL add cardiomegaly or lung opacity absent from the reference, while
our model reports only atelectasis.

\section{Prompt Example}
\label{app: instruction}
An example of the prompt used in our study is illustrated in the Prompt Example box below.


\begin{PromptBox}[float*=t, width=\textwidth]
{\large\bfseries\itshape \textless\textbar user\textbar\textgreater}

You are an expert radiologist. Your task is to analyze the given chest X-ray image, write a detailed finding, and then provide a structured summary of your finding.

You must follow this two-step process:
\begin{enumerate}[leftmargin=1.2em, itemsep=2pt, topsep=2pt]
  \item First, enclose your detailed, free-text radiology findings in \texttt{<think>} tags. This is your reasoning where you describe what you see.
  \item Second, based \textbf{ONLY} on the text you wrote in \texttt{<think>}, provide a structured JSON object in \texttt{<answer>} tags. The JSON must summarize your findings according to the 14 official CheXpert labels.
\end{enumerate}
CheXpert labels (14):
\begin{multicols}{2}
\begin{itemize}[leftmargin=1.2em, itemsep=0.5pt, topsep=0.5pt]
  \item Atelectasis
  \item Cardiomegaly
  \item Consolidation
  \item Edema
  \item Enlarged Cardiomediastinum
  \item Fracture
  \item Lung Lesion
  \item Lung Opacity
  \item No Finding
  \item Pleural Effusion
  \item Pleural Other
  \item Pneumonia
  \item Pneumothorax
  \item Support Devices
\end{itemize}
\end{multicols}

Use $1.0$ for positive findings, $0.0$ for negative, and $-1.0$ for uncertain.

\medskip
{\large\bfseries\itshape \textless\textbar assistant\textbar\textgreater}

{\normalsize\textbf{EXAMPLE:}}

\begin{tcolorbox}[enhanced, colback=white, colframe=gray!40, boxrule=0.4pt, arc=2pt,
  left=5pt,right=5pt,top=5pt,bottom=5pt]
\begin{lstlisting}[style=promptjson]
<think>
The cardiomediastinal silhouette is within normal limits. The lungs are clear.
There is no pleural effusion or pneumothorax. A port-a-cath is in place, with its
tip in the expected location.
</think>
<answer>
{
  "Atelectasis": 0.0,
  "Cardiomegaly": 0.0,
  "Consolidation": 0.0,
  "Edema": 0.0,
  "Enlarged Cardiomediastinum": 0.0,
  "Fracture": 0.0,
  "Lung Lesion": 0.0,
  "Lung Opacity": 0.0,
  "No Finding": 1.0,
  "Pleural Effusion": 0.0,
  "Pleural Other": 0.0,
  "Pneumonia": 0.0,
  "Pneumothorax": 0.0,
  "Support Devices": 1.0
}
</answer>
\end{lstlisting}
\end{tcolorbox}

\end{PromptBox}


\end{document}